%% file: main.tex
\documentclass[journal]{IEEEtran}

\usepackage{defs-common}
\usepackage{defs-custom}

\usepackage{algorithm}
\usepackage{algorithmic}
\usepackage{multirow}

\begin{document}

\title{Latent Combinational Game Design}

\author{Anurag Sarkar, Seth Cooper
\thanks{A. Sarkar and S. Cooper are both with the Khoury College of Computer Sciences, Northeastern University, Boston, MA, USA (email: sarkar.an@northeastern.edu, se.cooper@northeastern.edu)}
}

\markboth{IEEE Transactions on Games}%
{Shell \MakeLowercase{\textit{et al.}}: Bare Demo of IEEEtran.cls for IEEE Journals}

\maketitle
\begin{abstract}
We present \textit{latent combinational game design}---an approach for generating playable games that blend a given set of games in a desired combination using deep generative latent variable models.
We use Gaussian Mixture Variational Autoencoders (GMVAEs) which model the VAE latent space via a mixture of Gaussian components. Through supervised training, each component encodes levels from one game and lets us define blended games as linear combinations of these components. This enables generating new games that blend the input games as well as controlling the relative proportions of each game in the blend. We also extend prior blending work using conditional VAEs and compare against the GMVAE and additionally introduce a hybrid conditional GMVAE (CGMVAE) architecture which lets us generate whole blended levels and layouts.
Results show that \new{these} approaches can generate playable games that blend the input games in specified combinations.
We use both platformers and dungeon-based games to demonstrate our results.
\end{abstract}


\input{figuretable}
\input{body}

\bibliographystyle{IEEEtran}
\bibliography{refs-custom}
\clearpage

\end{document}

%% file: figuretable.tex

\newcommand{\XFIGURElcgd}{
\begin{figure*}[t!]
\centering
\includegraphics[width=0.7\textwidth]{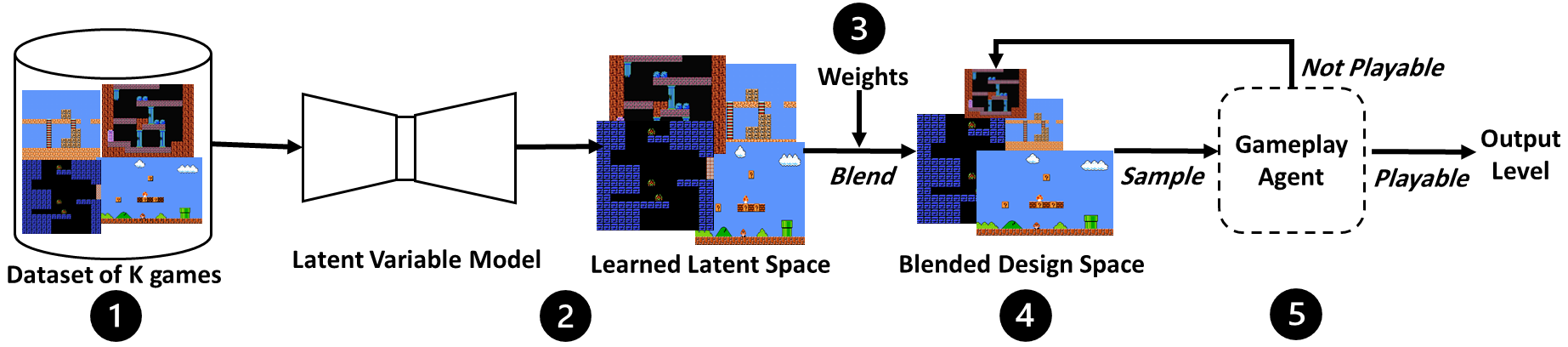}
\caption{\label{XFIGURElcgd} \small{\textit{Latent combinational game design:} 1) a dataset of k games for training 2) a model which learns a latent space on which 3) blend weights are applied to yield 4) a blended design space for sampling content satisfying 5) a constraints module such as a gameplay agent.}\vspace{-1.25em}}
\end{figure*}
}

\newcommand{\XFIGUREjump}{
\begin{figure}[t!]
\centering
\includegraphics[width=0.325\textwidth]{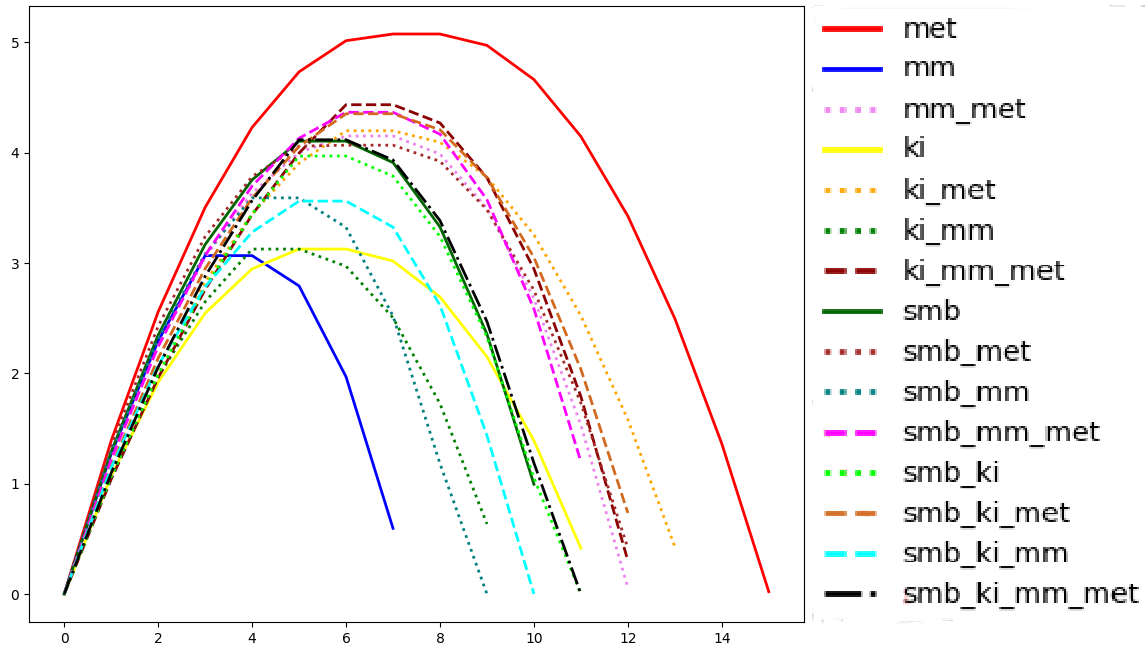}
\caption{\label{XFIGUREjump} \small{Jump arcs for various combinations of games.} \vspace{-1em}}
\end{figure}
}

\newcommand{\XFIGUREcgmvae}{
\begin{figure}[t!]
\centering
\includegraphics[width=0.4\textwidth]{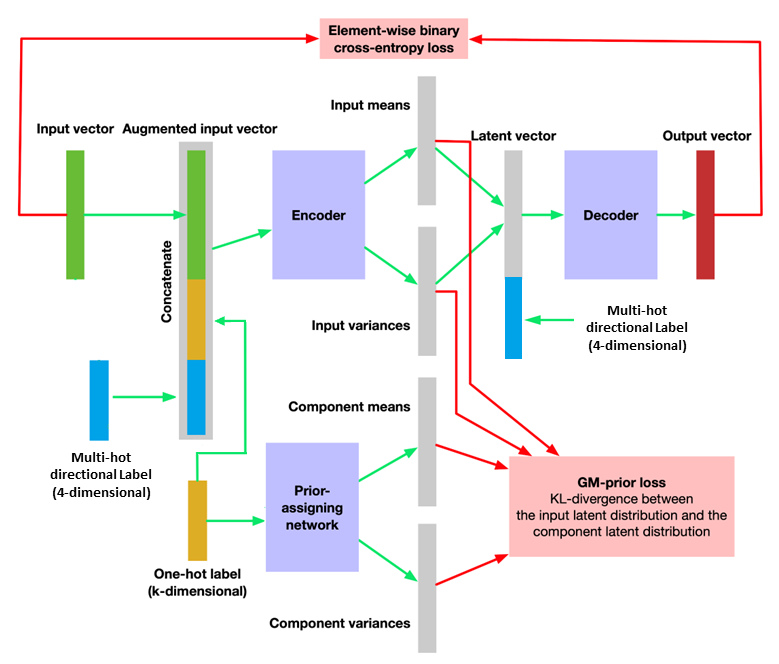}
\caption{\label{XFIGUREcgmvae} \small{Conditional GMVAE (CGMVAE) architecture.}}
\end{figure}
}

\newcommand{\XFIGUREdungeonegcc}{
\begin{figure}[t!]
\centering
\includegraphics[width=0.325\textwidth]{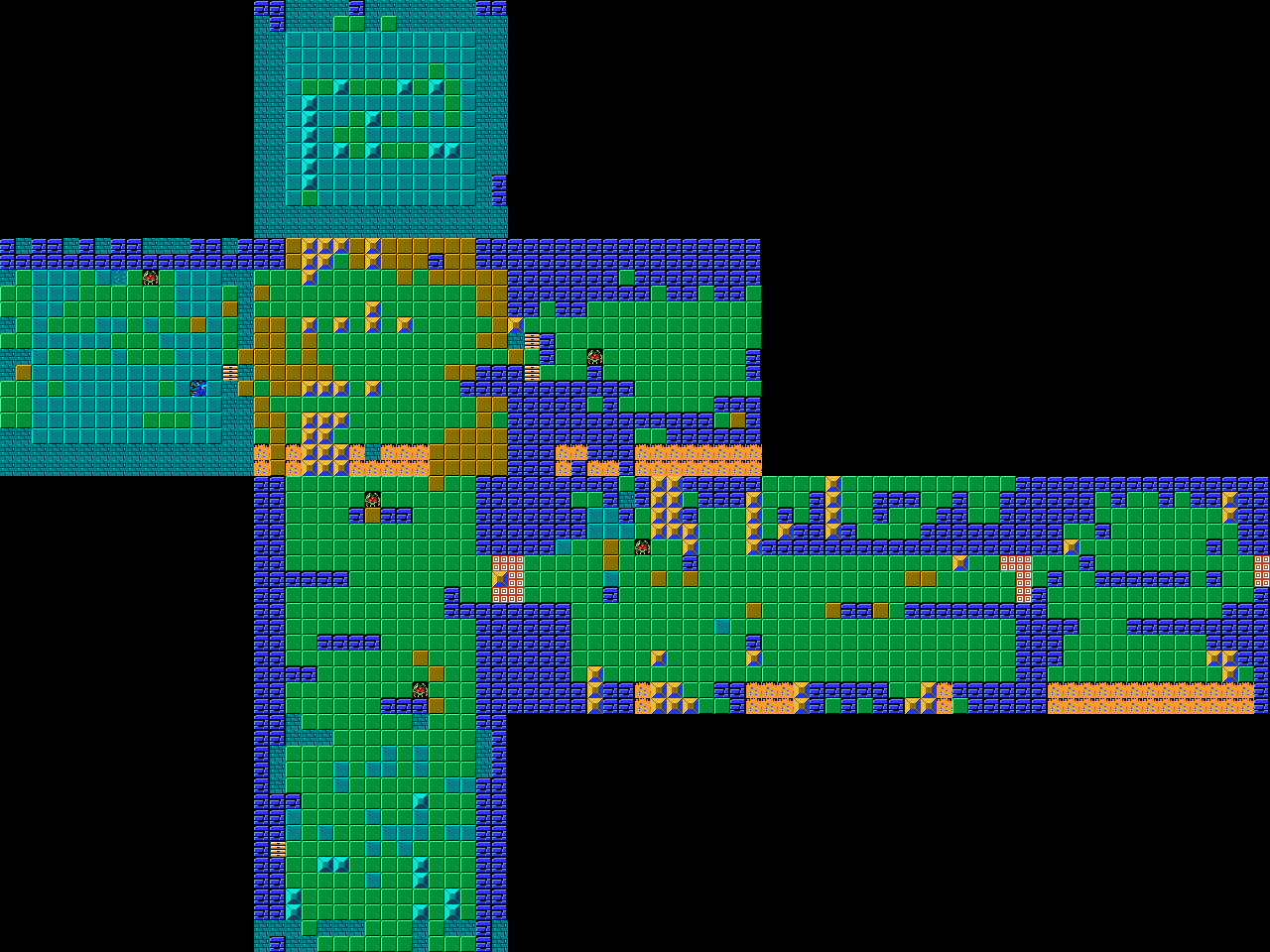}
\caption{\label{XFIGUREdungeonegcc} \small{Sample dungeon level generated using CCVAE-64} \vspace{-1.5em}}
\end{figure}
}

\newcommand{\XFIGUREdungeoneggm}{
\begin{figure}[t!]
\centering
\includegraphics[width=0.325\textwidth]{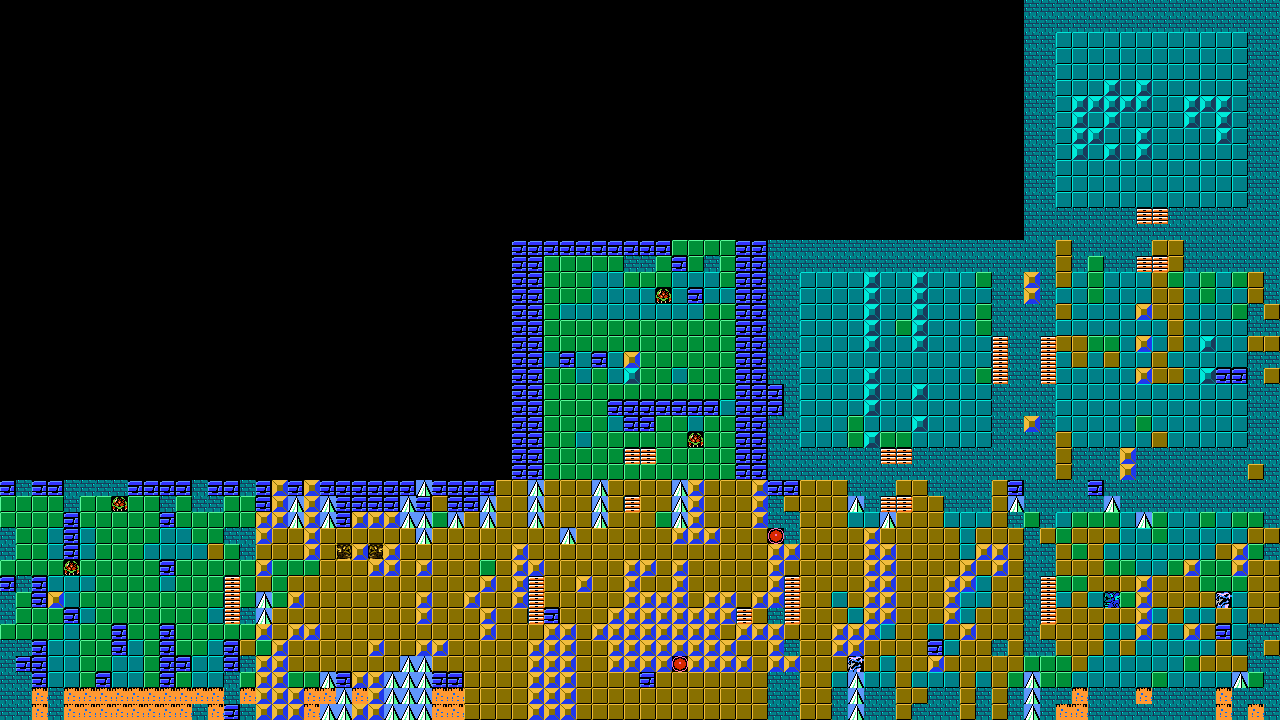}
\caption{\label{XFIGUREdungeoneggm} \small{Sample dungeon level generated using CGMVAE-32}\vspace{-1em}}
\end{figure}
}

\newcommand{\XFIGUREplatformeggm}{
\begin{figure}[t!]
\centering
\includegraphics[width=0.375\textwidth]{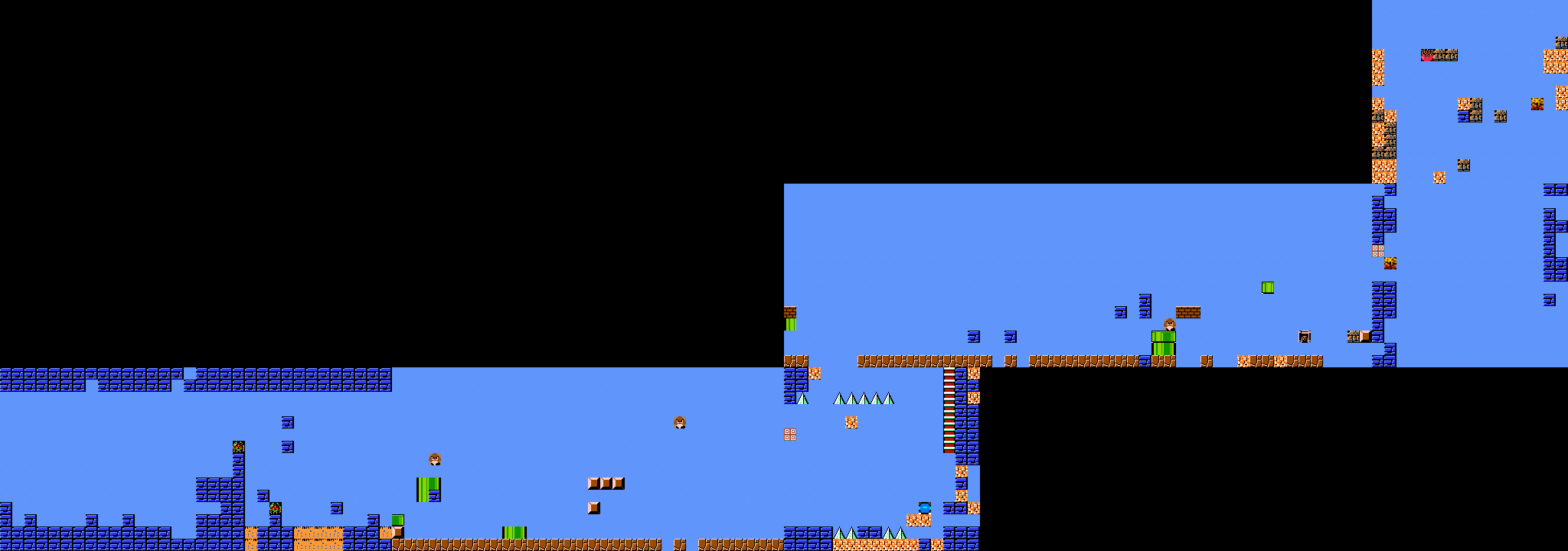}
\caption{\label{XFIGUREplatformeggm} \small{Sample platform level generated using CGMVAE-32}\vspace{-1em}}
\end{figure}
}

\newcommand{\XFIGUREplatformegcc}{
\begin{figure}[t!]
\centering
\includegraphics[width=0.375\textwidth]{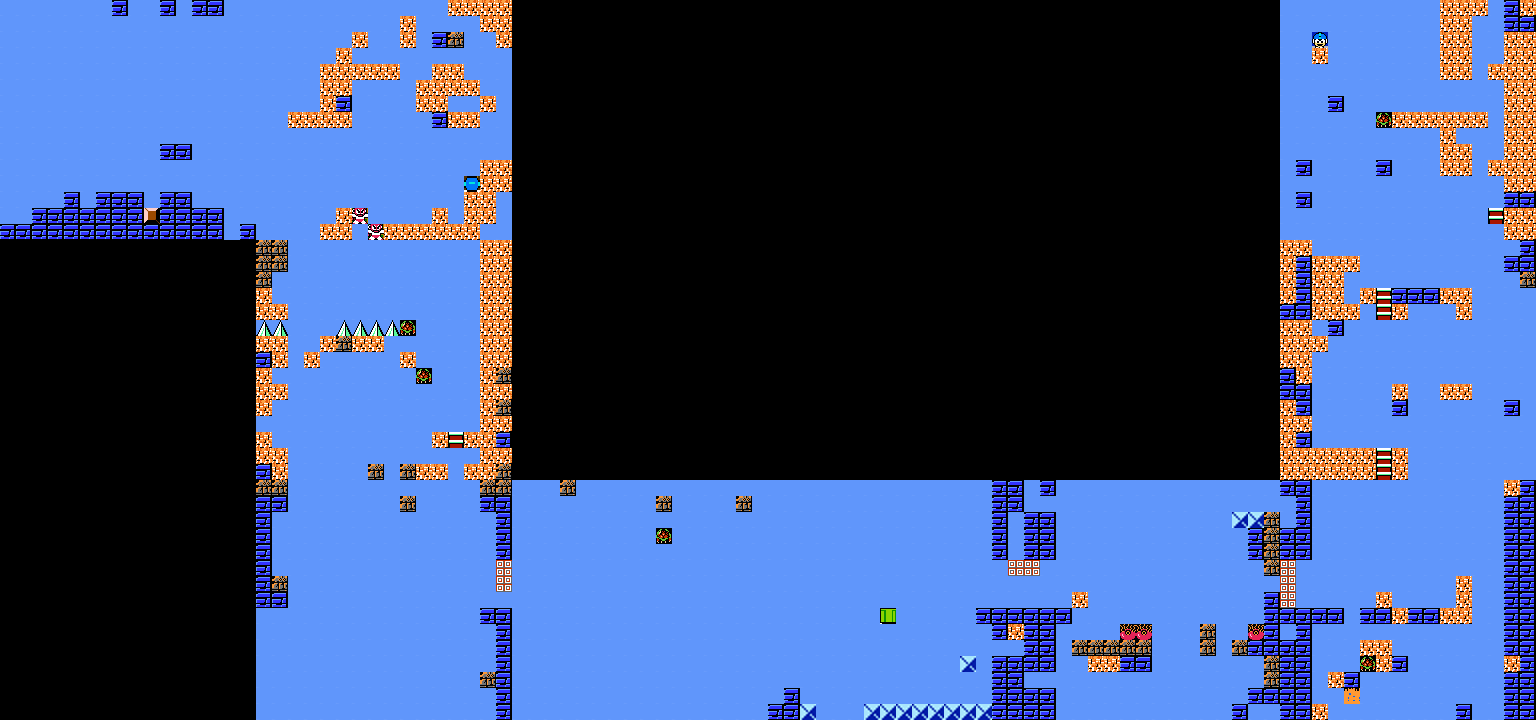}
\caption{\label{XFIGUREplatformegcc} \small{Sample platform level generated using CCVAE-64}\vspace{-1em}}
\end{figure}
}

\newcommand{\XTABLEdirlabels}{
\begin{table}[t]
\centering
\begin{tabular}{ccc}
\includegraphics[width=0.07\textwidth]{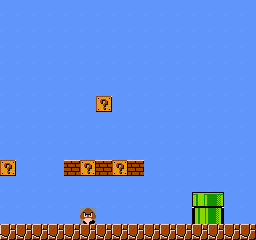} &
\includegraphics[width=0.07\textwidth]{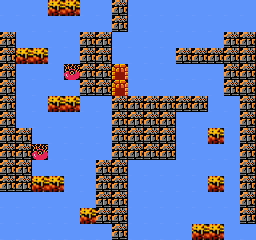} &
\includegraphics[width=0.07\textwidth]{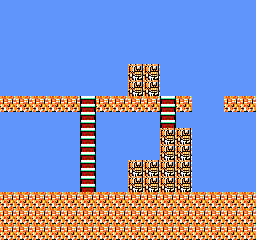} \\
Mario - \clabel{0,0,0,1} &
Kid Icarus - \clabel{1,1,0,0} &
Mega Man - \clabel{0,0,1,1} \\
\includegraphics[width=0.07\textwidth]
{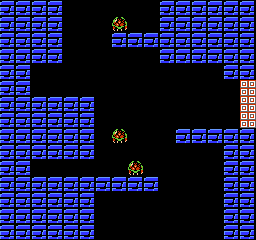} &
\includegraphics[width=0.09\textwidth]{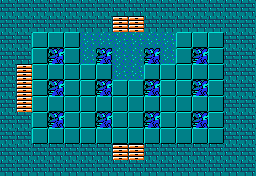} &
\includegraphics[width=0.07\textwidth]{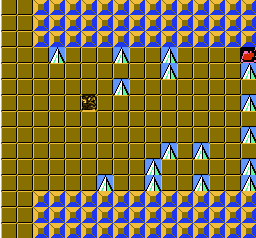}\\
Metroid - \clabel{1,1,0,1} &
Zelda - \clabel{1,1,1,0} &
DungeonGrams - \clabel{1,1,1,1}\\
\end{tabular}
\caption{\label{XTABLEdirlabels} \small{Example segments with corresponding directional labels indicating doors/openings in \clabel{Up, Down, Left, Right} directions.}\vspace{-1em}}
\end{table}
}

\newcommand{\XFIGUREjumps}{
\begin{figure}[h!]
\centering
\setlength{\tabcolsep}{2pt}
\begin{tabular}{cc}
\includegraphics[width=0.25\textwidth]{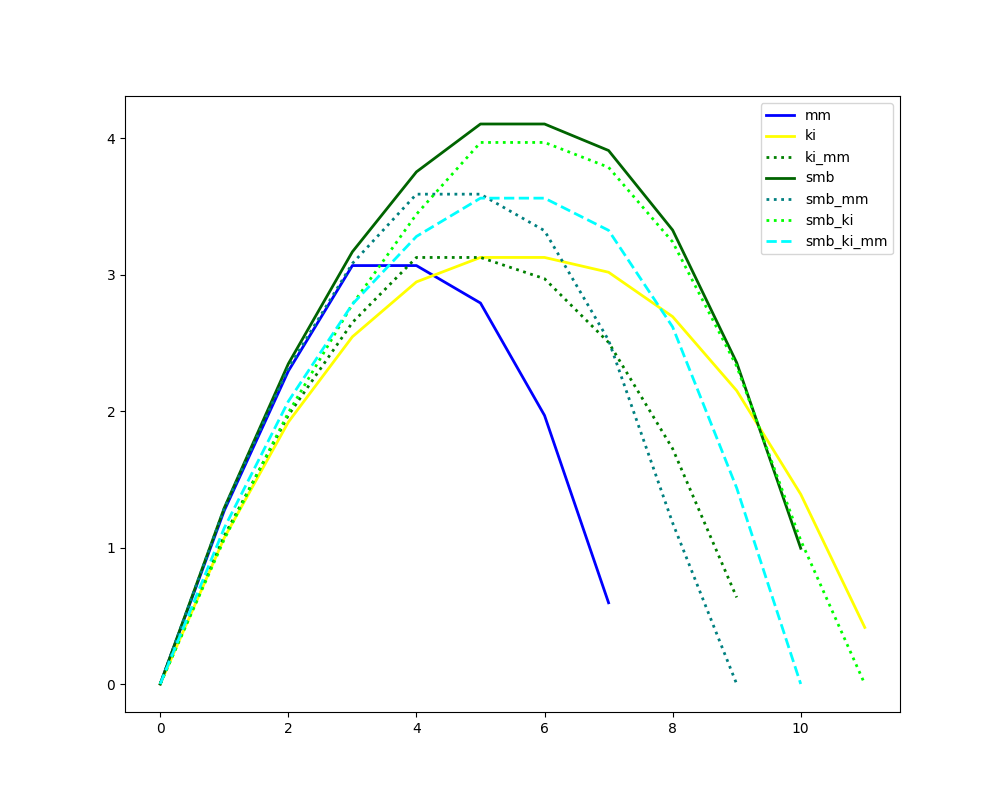}
&\includegraphics[width=0.25\textwidth]{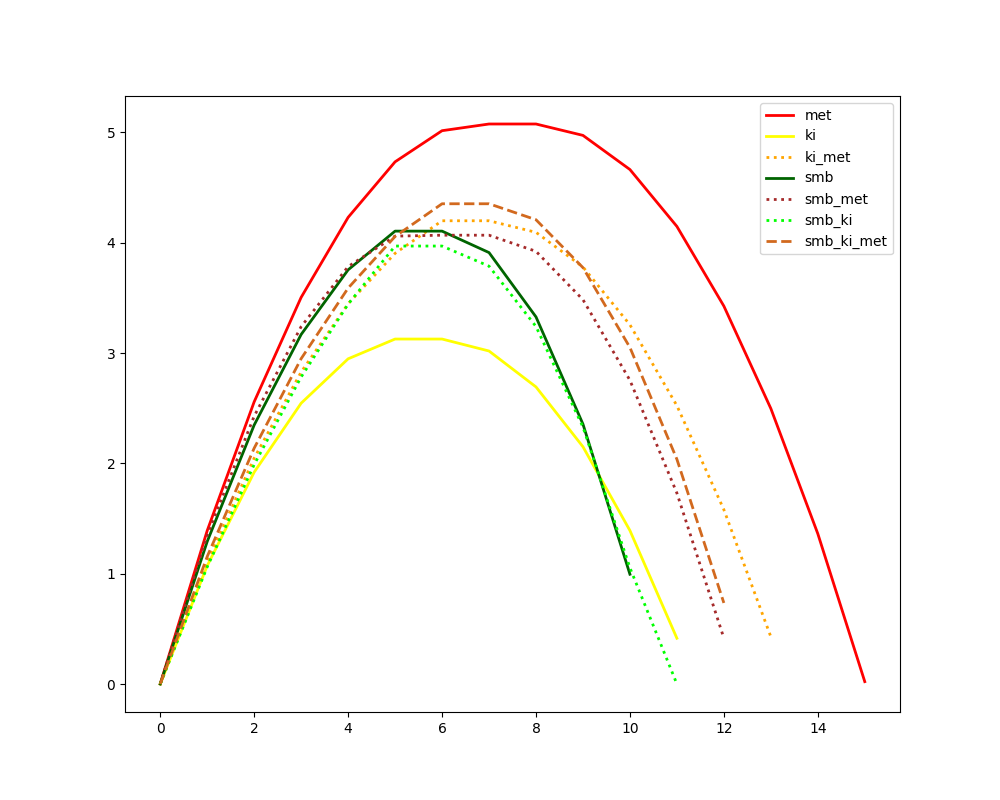}\\
SMB-KI-MM & SMB-KI-Met \\
\includegraphics[width=0.25\textwidth]{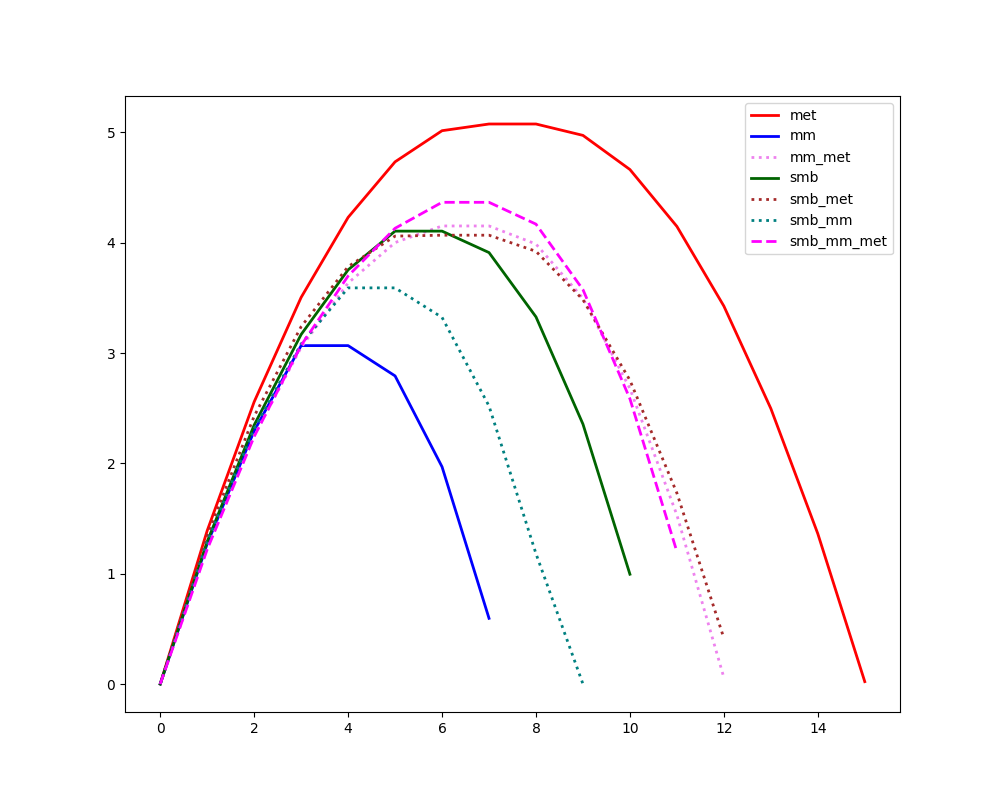}
&\includegraphics[width=0.25\textwidth]{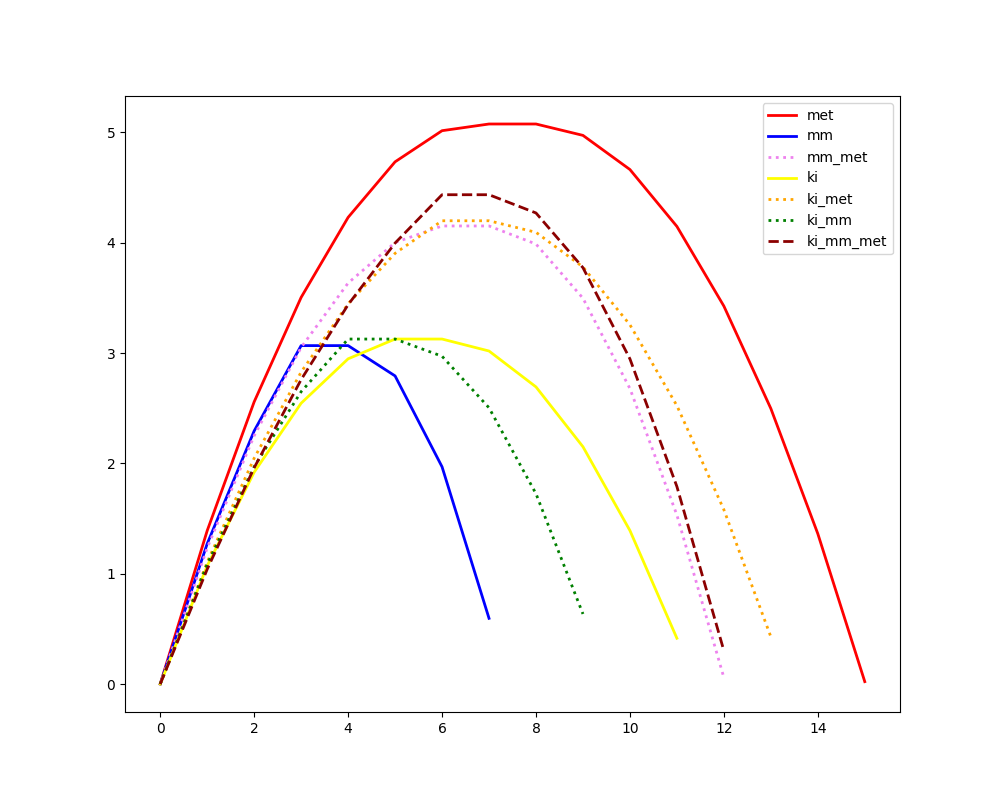}\\
SMB-MM-Met & KI-MM-Met \\
\end{tabular}
\caption{\label{XFIGUREjumps} Jump arcs for various combinations of games.}
\end{figure}
}

\newcommand{\XFIGUREgmvaeboth}{
\begin{figure}[t!]
\centering
\includegraphics[width=0.44\textwidth]{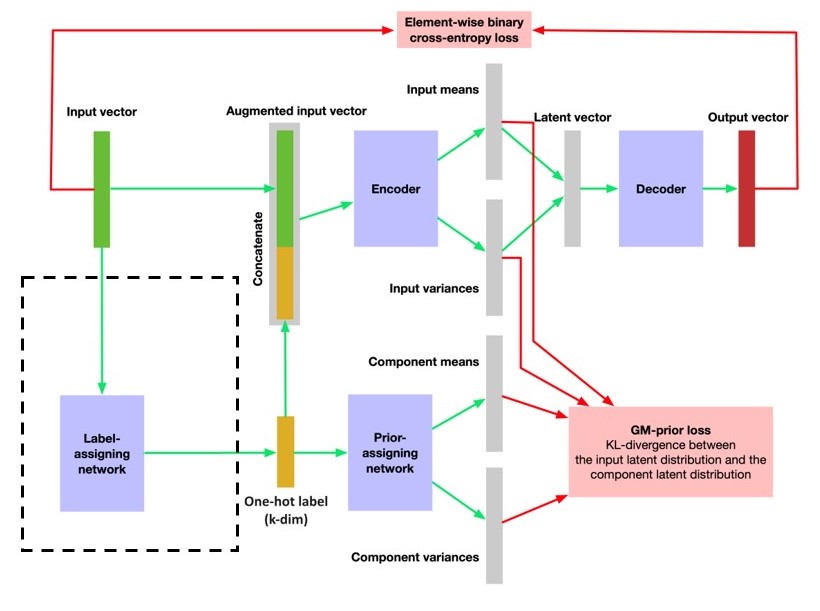}
\caption{\label{XFIGUREgmvaeboth} \small{\new{GMVAE architecture. The label-assigning network (dotted box) is part of the unsupervised GMVAE (as in prior work \protect\cite{yang2020game}) but not required in the supervised GMVAE setting used in this work.}} \vspace{-2em}}
\end{figure}
}

\newcommand{\XFIGUREexamples}{
\begin{figure}[t!]
\centering
\setlength{\tabcolsep}{1pt}
\scriptsize
\hspace*{-0.5cm}
\begin{tabular}{cccccc}
\raisebox{12pt}{\rotatebox{90}{\normalsize{\textbf{GM}}}}
&\includegraphics[width=0.08\textwidth]{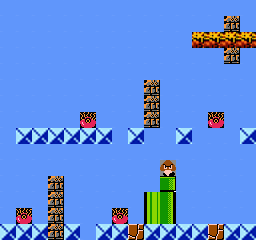}
&\includegraphics[width=0.08\textwidth]{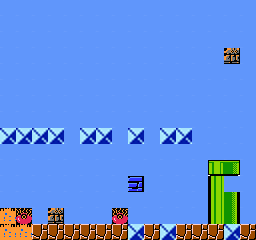}
&\includegraphics[width=0.08\textwidth]{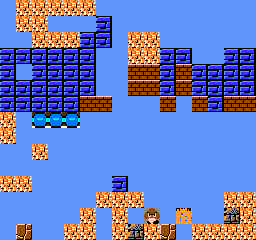}
&\includegraphics[width=0.08\textwidth]{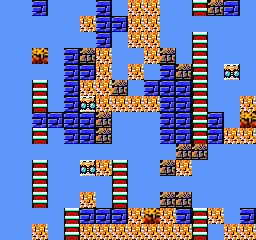}
&\includegraphics[width=0.08\textwidth]{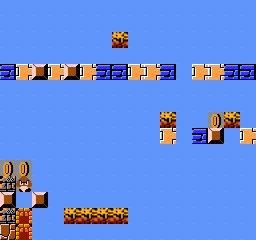}
\\
& \clabel{1110} & \clabel{1101}& \clabel{1011}& \clabel{0111} & \clabel{1111}\\
\raisebox{12pt}{\rotatebox{90}{\normalsize{\textbf{GM}}}}
&\includegraphics[width=0.08\textwidth]{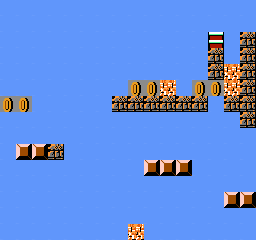}
&\includegraphics[width=0.08\textwidth]{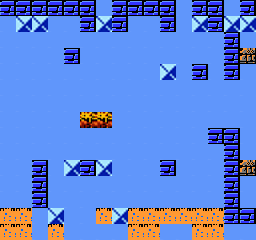}
&\includegraphics[width=0.08\textwidth]{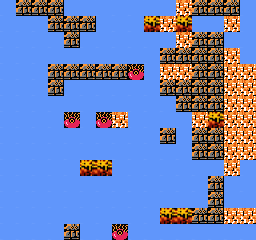}
&\includegraphics[width=0.08\textwidth]{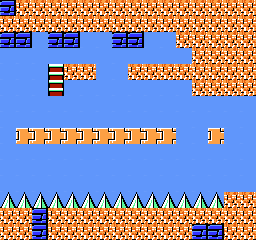}
&\includegraphics[width=0.08\textwidth]{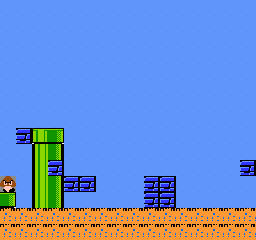}
\\
& \clabel{0.5,0.3,0.2,0.0} & \clabel{0.1,0.1,0.1,0.7}& \clabel{0.1,0.6,0.2,0.1}& \clabel{0,0.2,0.3,0.5} & \clabel{0.4,0,0,0.6}\\
\raisebox{6pt}{\rotatebox{90}{\normalsize{\textbf{CVAE}}}}
&\includegraphics[width=0.08\textwidth]{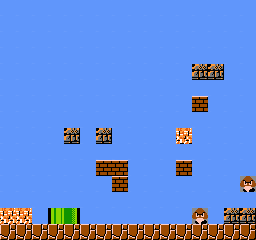}
&\includegraphics[width=0.08\textwidth]{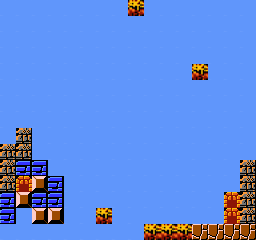}
&\includegraphics[width=0.08\textwidth]{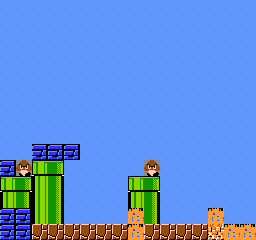}
&\includegraphics[width=0.08\textwidth]{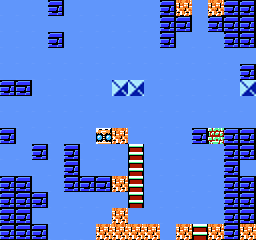}
&\includegraphics[width=0.08\textwidth]{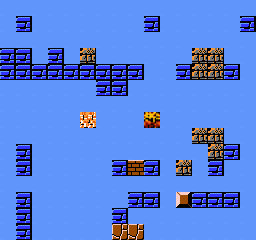}
\\
& \clabel{1110} & \clabel{1101}& \clabel{1011}& \clabel{0111} & \clabel{1111}\\
\raisebox{6pt}{\rotatebox{90}{\normalsize{\textbf{CVAE}}}}
&\includegraphics[width=0.08\textwidth]{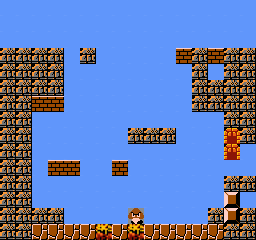}
&\includegraphics[width=0.08\textwidth]{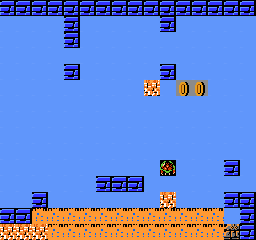}
&\includegraphics[width=0.08\textwidth]{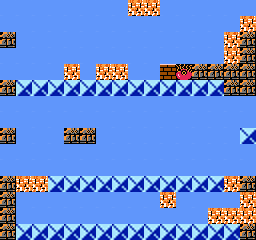}
&\includegraphics[width=0.08\textwidth]{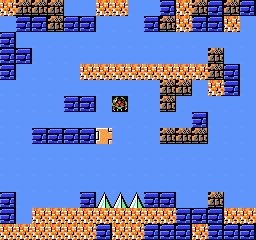}
&\includegraphics[width=0.08\textwidth]{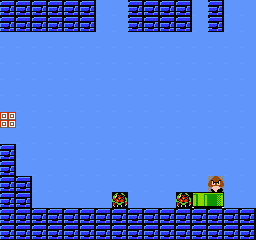}
\\
& \clabel{0.5,0.3,0.2,0.0} & \clabel{0.1,0.1,0.1,0.7}& \clabel{0.1,0.6,0.2,0.1}& \clabel{0,0.2,0.3,0.5} & \clabel{0.4,0,0,0.6}\\
\end{tabular}
\caption{\label{XFIGUREexamples} Example generated levels using the GM and CVAE models for a number of different blend weights. \vspace{-1.5em}}
\end{figure}
}


\newcommand{\XTABLEclassboth}{\begin{table}[t!]
\centering
\scriptsize
\setlength{\tabcolsep}{2pt}
\begin{tabular}{|c|c|c|c|c||c|c|c|c|}
\hline
 \multirow{2}{*}{} & \multicolumn{4}{c||}{Binary} & \multicolumn{4}{c|}{Fractional} \\
\hline
 & 8-dim & 16-dim & 32-dim & 64-dim & 8-dim & 16-dim & 32-dim & 64-dim \\
 \hline
CVAE & 1021.99 & 741.14 & 939.04 & \textbf{700.14} & 1113.8 & 995.08 & 1171.03 & \textbf{939.46} \\
GM & 659.83 & 707.9 & \textbf{293.79} & 310.94 & 904.21 & 937.26 & \textbf{851.84} & 947.31 \\
\hline
\end{tabular}
\caption{\label{XTABLEclassboth} \small{Classification results for blended levels generated using binary and fractional weights. Lower scores are better.} \vspace{-0.5em}}
\end{table}
}

\newcommand{\XTABLEblendagent}{\begin{table}[t!]
\scriptsize
\setlength{\tabcolsep}{2pt}
\centering
\begin{tabular}{|c|c|c|}
\hline
\clabel{SMB, KI, MM, Met} & GM-32 dim & CVAE-64 dim\\
\hline
 \clabel{0,0,0,1}	& 67.4 & 68.2\\
\clabel{0,0,1,0} &	49 & 48.3\\
\clabel{0,0,1,1} & 62.3 & 63.5\\
\clabel{0,1,0,0} &	69.3 & 68.9\\
\clabel{0,1,0,1}	& 52.2 & 70.1\\
\clabel{0,1,1,0}	& 52.7 & 72.2\\
\clabel{0,1,1,1}	& 53.5 & 76.9\\
\clabel{1,0,0,0}	& 83.8 & 89.8\\
\clabel{1,0,0,1}	& 69.1 & 71.1\\
\clabel{1,0,1,0} 	& 70.5 & 70.5\\
\clabel{1,0,1,1}	& 74.9 & 73.9\\
\clabel{1,1,0,0}	& 82.3 & 84.2\\
\clabel{1,1,0,1}	& 73.9 & 68.5\\
\clabel{1,1,1,0}	& 68.1 & 61.5\\
\clabel{1,1,1,1}	& 69.9 & 64.5\\
\hline
\end{tabular}
\caption{\label{XTABLEblendagent} \small{Percentage of playable levels for each binary blend.}}
\end{table}
}


\newcommand{\XTABLEblendagentfrac}{\begin{table}[t!]
\scriptsize
\setlength{\tabcolsep}{1pt}
\centering
\begin{tabular}{|c||c|cccc||c|cccc|}
\hline
 & \multicolumn{5}{c||}{GM-32 dim} & \multicolumn{5}{c|}{CVAE-64 dim} \\
 & Blend & SMB & KI & MM & Met & Blend & SMB & KI & MM & Met \\
 \hline
 \clabel{0.5, 0.3, 0.2, 0} &  75.3 & 67.3 & 8.5  & 6.5 & 0.2  &  77.9 & 66.4 & 21.8  & 4.6  & 2.2  \\
 \clabel{0.1, 0.1, 0.1, 0.7} & 62.1 & 0.2 & 0.6  & 0.7  & 66.4  & 61.8 & 0.5 & 0.3  & 1.2  & 64.8  \\
 \clabel{0.1, 0.6, 0.2, 0.1} & 71.3 & 9.9  & 58 & 6.9  & 1.5 & 71.9  & 3.1  & 63.2 & 8.2 & 3  \\
 \clabel{0, 0.2, 0.3, 0.5} & 53.6  & 0.7 & 3.7  & 9.7 & 51.3 & 60.5  & 0.6  & 2.5 & 14.7  & 58.9\\ 
 \hline
\end{tabular}
\caption{\label{XTABLEblendagentfrac} \small{Percentage of playable levels for fractional blends using the blended agent and original agents for each game.}}
\end{table}
}

\newcommand{\XTABLEtpklbinfull}{\begin{table}[t!]
\scriptsize
\setlength{\tabcolsep}{2pt}
\centering
\begin{tabular}{|c|cccc|cccc|}
\hline
 & \multicolumn{4}{c|}{GM-32 dim} & \multicolumn{4}{c|}{CVAE-64 dim} \\
 Blend Weights & SMB & KI & MM & Met & SMB & KI & MM & Met \\
 \hline
 \clabel{0001} & 12.68 & 13.02 & 12.83 & 1.59  &  13.51  & 12.97  & 13.16  & 1.37 \\
 \clabel{0010} & 10.95  & 10.64  & 2.1  & 11.28  & 9.87 & 10.23  & 1.64  & 11.03  \\
 \clabel{0011} & 11.28 & 10.69 & 7.59  & 5.59  &  8.7  & 8.78  & 6.37  & 5.2 \\
 \clabel{0100}& 10.77 & 2.6  & 10.55  & 11.7  & 11.02  & 1.62  & 10.97 & 11.61  \\
 \clabel{0101}& 10.84 & 8.57  & 11.07  & 4.46  & 9.04 & 8.16 & 9.16  & 2.96  \\
 \clabel{0110}& 8.58  & 6.74  & 5.34  & 10.03  & 7.95 & 4.79  & 5.14  & 8.52  \\
 \clabel{0111}&  9.01 & 7.61  & 5.78  & 6.35  & 6.74 & 6.65  & 5.15  & 4.26  \\
 \clabel{1000}& 0.8  & 4.67  & 4.31  & 4.49  & 0.65 & 4.43  & 4.26  & 4.64  \\
 \clabel{1001}& 6.76  & 7.73  & 7.49  & 2.36  &  5.03 & 5.84  & 5.78  & 1.9  \\
 \clabel{1010}& 3.53  & 6.15  & 2.51  & 6.4  & 2.94 & 4.09  & 1.81  & 4.49  \\
 \clabel{1011}& 5.31 & 6.29  & 4.25  & 4.42  & 3.95 & 4.56  & 3.18  & 3.22  \\
 \clabel{1100}& 3.6 & 4.25  & 5.97  & 6.51  & 2.44 & 2.48  & 4.12  & 4.06  \\
 \clabel{1101}& 4.78  & 4.06  & 6.1  & 4.32  & 2.58  & 3.03  & 3.13  & 1.49  \\
 \clabel{1110}& 4 & 5.29  & 3.94  & 6.56  & 2.19  & 2.02  & 2.46  & 3.54\\
 \clabel{1111}& 4.37  & 5.66  & 4.08  & 5.36  & 3.26  & 3.85  & 2.76  & 2.3 \\
 \hline
 \hline
 \clabel{0.5, 0.3, 0.2, 0} & 2.01 & 5 & 4.5 & 5.38  & 2.38  & 3.5  & 4.34  & 4.81 \\
 \clabel{0.1, 0.1, 0.1, 0.7} & 11.77  & 12.04  & 11.66  & 1.63  &   11.4 & 11.05  & 11.15  & 1.46  \\
 \clabel{0.1, 0.6, 0.2, 0.1} & 8.7 & 3.32 & 8.24  & 9.54  &  8.26  & 2.11  & 8.19  & 8.87 \\
 \clabel{0, 0.2, 0.3, 0.5} & 11.19 & 10.5  & 10.01  & 3.79  &  9.48  & 9.28  & 8.26 & 3.64  \\
 \hline
\end{tabular}
\caption{\label{XTABLEtpklbinfull} \small{TPKLDiv values between original games and the different blend configurations. Lower the value, closer the tile-pattern distribution of that blend is to the corresponding game.}}
\end{table}
}

\newcommand{\XTABLEtpklfracfull}{\begin{table}[t!]
\scriptsize
\setlength{\tabcolsep}{2pt}
\begin{tabular}{|c|cccc|cccc|cccc|}
\hline
 & \multicolumn{4}{c|}{GM1-32 dim} & \multicolumn{4}{c|}{GM2-64 dim} & \multicolumn{4}{c|}{CVAE-32 dim} \\
 Blend Weights & SMB & KI & MM & Met & SMB & KI & MM & Met & SMB & KI & MM & Met \\
 \hline
 \clabel{0.5, 0.3, 0.2, 0} & \textbf{2.01} & 5 & 4.5 & 5.38  & \textbf{1.49}  & 4.21  &  5.14 & 5.69  & \textbf{1.45}  & 3.29  & 3.65  & 4.33 \\
 \clabel{0.1, 0.1, 0.1, 0.7} & 11.77  & 12.04  & 11.66  & \textbf{1.63}  & 12.12  &  12.31 & 11.95  & \textbf{1.26}  &  11.16 & 10.85  & 11.06  & \textbf{1.4}  \\
 \clabel{0.1, 0.6, 0.2, 0.1} & 8.7 & \textbf{3.32} & 8.24  & 9.54  &  9.54 &  \textbf{1.65} & 9.88 & 10.49  & 7.65  & \textbf{2.03}  & 7.71  & 8.06 \\
 \clabel{0, 0.2, 0.3, 0.5} & 11.19 & 10.5  & 10.01  & \textbf{3.79}  & 11.08 & 11.52  & 11.34  & \textbf{1.76}  & 9.25  & 9.14  & 8.82 & \textbf{2.55}  \\
 \hline
 \end{tabular}
\caption{\label{XTABLEtpklfracfull} \small{TPKLDiv values between original games and the different fractional blend configurations. Lower the value, closer the tile-pattern distribution of that blend is to the corresponding game.}}
\end{table}
}

\newcommand{\XTABLEclassfull}{\begin{table}[t!]
\scriptsize
\setlength{\tabcolsep}{2pt}
\centering
\begin{tabular}{|c|ccccc|ccccc|}
\hline
& \multicolumn{5}{c|}{GM-32 dim} & \multicolumn{5}{c|}{CVAE-64 dim} \\
 Blend Weights & SMB & KI & MM & Met & Score & SMB & KI & MM & Met & Score\\
 \hline
 \clabel{0001} & 0.7 & 0.1 & 0.4 & 98.8 & 2.1  & 0.1  & 0  & 0  & 99.9 & 0.02\\
 \clabel{0010} & 5.2  & 1.8  & 93.0  & 0 & 79.3 & 1.5 & 0.9  & 97.6  & 0  & 8.8\\
 \clabel{0011} & 4.4 & 1.3 & 40.5  & 53.8 & 125.7  &  6.2  & 0  & 58.4  & 35.4 & 322.2\\
 \clabel{0100}& 2.3 & 96.5  & 1  & 0.2 & 18.6 & 0.4  & 99.6  & 0 & 0  & 0.3\\
 \clabel{0101}& 11.7 & 21.1  & 2.1  & 65.1 & 1204.5 & 17.7 & 5.5 & 0.6  & 76.2 & 2980.3\\
 \clabel{0110}& 11.8  & 38  & 50  & 0.2 & 283.3 & 23.8 & 45.7 & 30.5  & 0 & 965.2 \\
 \clabel{0111}&  15.6 & 20.4  & 34.1  & 29.9  & 422 & 35.5 & 7.5  & 30.9 & 26.1  & 1983.5\\
 \clabel{1000}& 94.6  & 3.8  & 1.6  & 0 & 46.2 & 98.6 & 1.4  & 0  & 0  & 3.9\\
 \clabel{1001}& 34.8  & 2.1  & 2.3  & 60.8 & 357.4  & 39.9 & 1.4 & 0 & 58.7  & 179.7\\
 \clabel{1010}& 48.1  & 5.3  & 46.6  & 0 & 43.3 & 42.7 & 9.3 & 48 & 0  & 143.8\\
 \clabel{1011}& 39 & 2.8  & 38.8  & 19.4 & 263.8 & 40.8 & 6.1 & 36.9 & 16.2 & 398.8\\
 \clabel{1100}& 56 & 42.8  & 1.2  & 0 & 89.3 & 64.5 & 35.5 & 0 & 0 & 420.5\\
 \clabel{1101}& 50.6  & 25.7  & 3  & 20.7 & 524.8  & 55.9 & 22.1  & 0.2 & 21.8 & 768.5\\
 \clabel{1110}& 47.8 & 26.4  & 25.7  & 0.1  & 315.6 & 30.2 & 62.6 & 7.2  & 0 & 1549.3\\
 \clabel{1111} & 42.5  & 19.4  & 29.6  & 8.5  & 631 & 40.5 & 37.1  & 13.4 & 9 & 777.2\\
 \hline
  \hline
 \clabel{0.5, 0.3, 0.2, 0} & 74.4 & 18.6 & 7 & 0 & 894.3 & 65.9  & 33.1  & 1  & 0 & 623.4\\
 \clabel{0.1, 0.1, 0.1, 0.7} & 2.3  & 0.7  & 1.8  & 95.2 & 848.1 &  2.4 & 0.2  & 0.5  & 96.9 & 967.7\\
 \clabel{0.1, 0.6, 0.2, 0.1} & 12.6 & 77.3 & 9.3  & 0.8 & 505.2 &  9.8  & 88.9 & 1.2  & 0.1 & 1286.7\\
 \clabel{0, 0.2, 0.3, 0.5} & 8.3 & 3.1  & 14.7  & 73.9 & 1159.8 & 12.3  & 2.5  & 18.3 & 66.9  & 880\\
 \hline
\end{tabular}
\caption{\label{XTABLEclassfull} \small{Full classification results for both blend weight types. Lower the score, more aligned are predictions with weights.}}
\end{table}
}

\newcommand{\XTABLEclassfullplt}{\begin{table}[h!]
\scriptsize
\setlength{\tabcolsep}{2pt} \centering 
\begin{tabular}{|c|ccccc|ccccc|}
\hline
& \multicolumn{5}{c|}{CGM-32 dim} & \multicolumn{5}{c|}{CCVAE-64 dim} \\
 Blend Weights & SMB & KI & MM & Met & Score & SMB & KI & MM & Met & Score\\
 \hline
 \clabel{0001} & 4.7 & 4 & 0.6 & 90.7 & 124.9 & 0.4  & 0  & 0.1  & 99.5 & 0.4\\
 \clabel{0010} & 6.5  & 10.6  & 82.8  & 0.1 & 450.5 & 7 & 4  & 89  & 0  & 186\\
 \clabel{0011} & 12.5 & 7.1 & 45  & 35.4 & 444.8 &  8.1  & 0.2  & 18.4  & 73.3 & 1607.1\\
 \clabel{0100}& 1.9 & 97.3  & 0.6  & 0.2 & 11.3  & 0.1  & 98  & 1.9 & 0  & 7.6\\
 \clabel{0101}& 13.9 & 33.6  & 0.9  & 51.6  & 465.5 & 22.5 & 44.2 & 0.1  & 33.2 & 822.1\\
 \clabel{0110}& 13  & 40.3  & 46.5  & 0.2 & 275.4  & 25.4 & 32.8 & 41.8  & 0 & 1008.2\\
 \clabel{0111}&  15.2 & 26.7  & 28.1  & 30 & 312.5  & 29.8 & 52.3  & 15.6 & 2.3 & 2523.3 \\
 \clabel{1000}& 53.1  & 46.1  & 0.8  & 1 & 4326.5 & 92.5 & 7.5  & 0.1  & 0  & 112.5\\
 \clabel{1001}& 35.9  & 8.6  & 0.6  & 54.9  & 297.1 & 8.6 & 1.1 & 0.7 & 90.3 & 3339.8\\
 \clabel{1010}& 31.7  & 13.3  & 54.8  & 0.2  & 534.9 & 62.1 & 22.5 & 15.4 & 0  & 1849.8 \\
 \clabel{1011}& 33.8 & 6.9  & 35.5  & 23.8  & 143 & 50.2 & 12.3 & 16.7 & 20.8 & 868.7\\
 \clabel{1100}& 23.5 & 75.4  & 0.8  & 0.3 & 1348.1 & 58.1 & 41.9 & 0 & 0 & 131.2\\
 \clabel{1101}& 36.5  & 34.3  & 0.8  & 28.4  & 35.9 & 19.5 & 75.2  & 0.6 & 4.7 & 2764.4\\
 \clabel{1110}& 33.2 & 37.1  & 29  & 0.7 & 33.5 & 22 & 75.6 & 2.4  & 0 & 2871.8\\
 \clabel{1111}& 36.1  & 20.6  & 24.4  & 18.9 & 500.1 & 36 & 47  & 7.5 & 9.5 & 1151.5\\
 \hline
 \hline
 \clabel{0.5, 0.3, 0.2, 0} & 10.7 & 88.7 & 0.5 & 0.1 & 2106.6 & 28.8  & 71.1 & 0.1  & 0.2 & 1221\\
 \clabel{0.1, 0.1, 0.1, 0.7} & 15.1  & 14.4  & 0.5  & 70 & 540.7 &  1.6 & 0.2  & 0.1  & 98.2 & 803.8\\
 \clabel{0.1, 0.6, 0.2, 0.1} & 2.8 & 95.9 & 1.2  & 0.1  & 1516.3 &  0.6  & 99.4 & 5.3  & 0.2 & 974.5\\
 \clabel{0, 0.2, 0.3, 0.5} & 21.6 & 43.9  & 3.5  & 31 & 1355.9 & 11  & 4.2  & 0.4 & 84.4  & 1752.8\\
 \hline
\end{tabular}
\caption{\label{XTABLEclassfullplt} \small{Full classification results on platformers. Lower the score, more aligned are predictions with weights.}\vspace{-0.5em}}
\end{table}
}

\newcommand{\XTABLEclassfulldg}{\begin{table}[h!]
\scriptsize
\setlength{\tabcolsep}{2pt} \centering 
\begin{tabular}{|c|cccc|cccc|}
\hline
& \multicolumn{4}{c|}{CGM-32 dim} & \multicolumn{4}{c|}{CCVAE-64 dim} \\
 Blend Weights & Zelda & DGG & Met & Score & Zelda & DGG & Met & Score \\
 \hline
 \clabel{001} & 0.2 & 0.3 & 99.7 & 0.2 & 0 & 0 & 100 & 0\\
 \clabel{010} & 0.1  & 99.8  & 0.1 & 0.1 & 0.1 & 99.8 & 0.1 & 0.1\\
 \clabel{011} & 0.3 & 57.8  & 41.9 & 126.5 & 0.7 & 9.9 & 89.4 & 3160.9\\
 \clabel{100}& 100 & 0 & 0 & 0 & 1.8 & 70.1 & 28.1 & 15346.9\\
 \clabel{101}& 74.3 & 1.8 & 23.9 & 1275 & 0.1 & 0.2 & 99.7 & 4960.1\\
 \clabel{110}& 42.1  & 57.6  & 0.3 & 120.3 & 0.2 & 99.4 & 0.4 & 4920.6\\
 \clabel{111}&  26  & 46.2  & 27.8 & 250 & 2.4 & 61.1 & 36.5 & 1737.9\\
 \hline
 \hline
 \clabel{0.5, 0.3, 0.2} & 39.5 & 59.3 & 1.2 & 1322.2 & 0.6 & 68.5 & 30.9 & 4041.4\\
 \clabel{0.2, 0.5, 0.3} & 0.4  & 95.3  & 4.3 & 1642.6  & 1 & 78.1 & 20.9 & 1233.4\\
 \clabel{0.3, 0.2, 0.5} & 3.7 & 31.4 & 64.9 & 1043.7 & 0.1 & 0.1 & 99.9 & 3770.1\\
 \hline
\end{tabular}
\caption{\label{XTABLEclassfulldg} \small{Full classification results on dungeon games.Lower the score, more aligned are predictions with weights.} \vspace{-0.25em}}
\end{table}
}

\newcommand{\XTABLEtpklbinfullplt}{\begin{table}[t!]
\scriptsize
\setlength{\tabcolsep}{2pt} \centering 
\begin{tabular}{|c|cccc|cccc|} \hline
 & \multicolumn{4}{c|}{CGM-32 dim} & \multicolumn{4}{c|}{CCVAE-64 dim} \\
 Blend Weights & SMB & KI & MM & Met & SMB & KI & MM & Met \\
 \hline
 \clabel{0001} & 12.67 & 12.35 & 12.65 & 2.22  &  13.16  & 12.92  & 13.2  & 1.51 \\
 \clabel{0010} & 10.67  & 10.49  & 2.39 & 11.55  & 11.1 & 10.99  & 1.99  & 11.92  \\
 \clabel{0011} & 10.39 & 10.23 & 6.61  & 6.94  &  8.52  & 8.61  & 5.55  & 5.69 \\
 \clabel{0100}& 9.11 & 3.06  & 9.37  & 10.03  & 10.82  & 2  & 10.86 & 11.67  \\
 \clabel{0101}& 10.58 & 8.95  & 10.67  & 4.28  & 8.63 & 6.95 & 8.74  & 4.07  \\
 \clabel{0110}& 9.27  & 7.55  & 4.29  & 10.17  & 9.54 & 7.64  & 4.41 & 10.35  \\
 \clabel{0111}&  9.36 & 8.42  & 6.77  & 6.62  & 5.72 & 5.33  & 3.56 & 5.18  \\
 \clabel{1000}& 2.88  & 4.1  & 5.06  & 5.68  & 0.94  & 5.26  & 5.12  & 5.7 \\
 \clabel{1001}& 8  & 8.53  & 8.75  & 2.75  &  6.64 & 7.12  & 7.1  & 1.59  \\
 \clabel{1010}& 6.12  & 6.68  & 2.29  & 7.42  & 5.17 & 6.06  & 1.97  & 6.53  \\
 \clabel{1011}& 6.83 & 7.34  & 5.07 & 4.98  & 4.52 & 4.99  & 2.92  & 3.6  \\
 \clabel{1100}& 5.04 & 3.21  & 6.43  & 7.05  & 3.3 & 2.52  & 4.88  & 5.36  \\
 \clabel{1101}& 6.94  & 6.52  & 7.82  & 4.14  & 3.45  & 2.97  & 3.88  & 2.62 \\
 \clabel{1110}& 5.8 & 5.46  & 3.54 & 7.22  & 4.5  & 3.65  & 3.02 & 5.5\\
 \clabel{1111}& 6.33  & 6.39  & 5.16  & 5.27  & 3.86  & 3.97  & 2.92  & 3.21 \\
 \hline
 \hline
 \clabel{0.5, 0.3, 0.2, 0} & 4.76 & 3.77 & 5.4 & 6.68  & 3.7  & 3.89  & 4.42  & 5.86 \\
 \clabel{0.1, 0.1, 0.1, 0.7} & 11.32  & 10.98  & 11.35  & 2.53  & 11.23 & 11.12  & 11.27  & 1.65  \\
 \clabel{0.1, 0.6, 0.2, 0.1} & 7.46 & 3.74 & 7.42  & 8.38  &  7.92  & 3.08  & 7.05  & 8.75 \\
 \clabel{0, 0.2, 0.3, 0.5} & 9.42 & 8.82  & 8.65  & 4.29  &  8.88  & 8.81  & 7.77 & 3.62  \\
 \hline
\end{tabular}
\caption{\label{XTABLEtpklbinfullplt} \small{TPKLDiv values between each original platformer and the different blend configurations. Lower the value, closer the tile-pattern distribution of that blend is to the corresponding game.}}
\end{table}
}

\newcommand{\XTABLEtpklbinfulldg}{\begin{table}[t!]
\scriptsize
\setlength{\tabcolsep}{2pt} \centering 
\begin{tabular}{|c|ccc|ccc|} \hline
 & \multicolumn{3}{c|}{CGM-32 dim} & \multicolumn{3}{c|}{CCVAE-64 dim} \\
 Blend Weights & Zelda & DGG & Met & Zelda & DGG & Met \\
 \hline
 \clabel{001} & 23.37 & 23.4 & 1.77  &  23.64  & 23.68  & 1.38 \\
 \clabel{010} & 21.92  & 3.73 & 21.78  & 21.85 & 3.96  & 21.71  \\
 \clabel{011} & 20.45 & 13.94 & 12.52  & 21.03  &  19.95  & 8.3 \\
 \clabel{100}& 1.28  & 22.8  & 22.62  & 10.87  & 18.73 & 18.69  \\
 \clabel{101}& 8.35  & 20.79 & 17.61 & 23.16  & 23.27 & 2.76  \\
 \clabel{110}& 10.45  & 15.5  & 20.07  & 16.42 & 12.07  & 19.84  \\
 \clabel{111}& 14.48  & 15.72 & 15.68  & 16.45  & 17.82 & 15.83  \\
 \hline
 \hline
 \clabel{0.5, 0.3, 0.2} & 7.17 & 19.1 & 20.3  & 16.91  & 16.71  & 15.39 \\
 \clabel{0.2, 0.5, 0.3} & 20.29  & 7.87 & 19.04  & 18.89  & 14.47 & 15.53  \\
 \clabel{0.3, 0.2, 0.5} & 19.83 & 20.1 & 7.9 & 22.47  & 22.54  & 3.83\\
 \hline
\end{tabular}
\caption{\label{XTABLEtpklbinfulldg} \small{TPKLDiv values between each original dungeon game and the different blend configurations. Lower the value, closer the tile-pattern distribution of that blend is to the corresponding game.}}
\end{table}
}

\newcommand{\XTABLEcgmccclass}{\begin{table}[t!]
\scriptsize
\centering
\begin{tabular}{|c|cc|cc|}
\hline
\multirow{2}{*}{} & \multicolumn{2}{c|}{Platformers}          & \multicolumn{2}{c|}{Dungeons}             \\ \cline{2-5} 
                  & \multicolumn{1}{c|}{Binary}  & Fractional & \multicolumn{1}{c|}{Binary}  & Fractional \\ \hline
CGM-32            & \multicolumn{1}{c|}{\textbf{620.27}}  & 1379.86    & \multicolumn{1}{c|}{\textbf{253.14}}  & \textbf{1820.86}    \\ 
CC-64             & \multicolumn{1}{c|}{1282.97} & \textbf{1188.01}    & \multicolumn{1}{c|}{4303.77} & 3014.97    \\ \hline
\end{tabular}
\caption{\label{XTABLEcgmccclass} \small{Classification results for blended levels generated by the CGM-32 and CC-64 models. Lower scores are better.}}
\end{table}
}

\newcommand{\XTABLEdirsummboth}{\begin{table}[t!]
\scriptsize
\centering
\setlength{\tabcolsep}{3pt}
\begin{tabular}{|c|c|ccc|ccc|}
\hline
\multirow{2}{*}{\begin{tabular}[c]{@{}c@{}}Genre\end{tabular}} & \multirow{2}{*}{Model} & \multicolumn{3}{c|}{Binary} & \multicolumn{3}{c|}{Fractional} \\ \cline{3-8} 
 &  & \multicolumn{1}{c|}{Exact} & \multicolumn{1}{c|}{Adm} & Inadm & \multicolumn{1}{c|}{Exact} & \multicolumn{1}{c|}{Adm} & \multicolumn{1}{l|}{Inadm} \\ \hline
\multirow{2}{*}{Platformers} & CGM-32 & \multicolumn{1}{c|}{23.03} & \multicolumn{1}{c|}{63.4} & 36.6 & \multicolumn{1}{c|}{33.34} & \multicolumn{1}{c|}{84.41} & 15.59 \\ 
 & CC-64 & \multicolumn{1}{c|}{27.05} & \multicolumn{1}{c|}{71.01} & 28.99 & \multicolumn{1}{c|}{36.24} & \multicolumn{1}{c|}{82.28} & 17.72 \\ \hline
\multirow{2}{*}{Dungeons} & CGM-32 & \multicolumn{1}{c|}{47.33} & \multicolumn{1}{c|}{83.77} & 16.23 & \multicolumn{1}{c|}{49.67} & \multicolumn{1}{c|}{96.29} & 3.71 \\ 
 & CC-64 & \multicolumn{1}{c|}{33.43} & \multicolumn{1}{c|}{79.31} & 20.69 & \multicolumn{1}{c|}{36.77} & \multicolumn{1}{c|}{83.54} & 16.45 \\ \hline
\end{tabular}
\caption{\label{XTABLEdirsummboth} \small{Percentage of exact, admissible (adm) and inadmissible (inadm) matches for models using both weight types.}\vspace{-1em}}
\end{table}
}

%% file: body.tex
\vspace{-0.5em}
\section{Introduction}
Methods for Procedural Content Generation via Machine Learning (PCGML) \cite{summerville2017procedural} primarily focus on learning distributions of individual games, but an emerging body of work \cite{sarkar2023pcgkt} has focused on methods for recombining learned models and distributions across multiple games to generate content for entirely new games. These have been motivated by limited training data, wanting to leverage design knowledge in one set of games to apply in another and exploring creative PCGML applications beyond level generation. One such application is game blending---the generation of new games by blending the levels and/or mechanics of existing games, \new{inspired by designers building new games by combining ideas from existing ones \cite{gow2015towards}}. 
An extensively used approach for blending levels has been the variational autoencoder (VAE) \cite{kingma2013autoencoding}, an encoder-decoder architecture that learns latent representations of data. Prior works  \cite{sarkar2019blending,sarkar2020exploring} have demonstrated VAEs to be capable of such blending but offer limited means to control the blending, \new{learning a combined latent space spanning all games without affording the ability to sample specific blends or games.}
Given a design space spanning a set of games, one may wish to blend only a certain subset of the games and in desired proportions relative to one another. To this end, prior works \cite{sarkar2020conditional,sarkar2021dungeon} have shown via conditional VAEs that supervision via labels is promising for obtaining such desired blend combinations. However, these works only examined binary combinations (a game is either included or not) rather than relative blend ratios.  
Further, the playability of such blended levels has been previously evaluated only by using mechanics for the original games as proxies, rather than by blending the mechanics themselves.

\XFIGURElcgd

In this work, we introduce a framework for generating playable blended games that blend desired proportions of games. We call this \textit{latent combinational game design}---\textit{latent} since we use latent representations, \textit{combinational} since game blending is a combinational creativity \cite{boden2004creative} process and \textit{game design} since we generate novel, playable games. Overall, the framework consists of:
\begin{itemize}
    \item Training a latent variable model to learn a design space that captures the set of $k$ games to be blended
    \item Defining $k$ weights specifying blend proportions
    \item Using the weights to generate blended levels by sampling from the relevant parts of the learned design space
    \item Using the weights to derive a blended agent that combines the mechanics of the original game agents
\end{itemize}
We apply this framework using supervised Gaussian Mixture VAEs (GMVAE)  \cite{dilokthanakul2016deep,shu2016gaussian}
and conditional VAEs (CVAE) \cite{sohn2015learning} which we train using levels from the platformers \textit{Super Mario Bros.}, \textit{Kid Icarus}, \textit{Mega Man} and \textit{Metroid}. Moreover, unlike prior approaches, we test playability using blended agents derived by combining the jump physics of these games using the same weights used to blend the levels. Overall, we find that \new{the framework (with both models) enables} generating playable blended games based on the weights. 
Further, since these models generate segments, we introduce a novel hybrid conditional GMVAE (CGMVAE) architecture which enables generating whole blended levels as well as layouts that blend \new{levels of the} dungeon games \textit{The Legend of Zelda}, \textit{DungeonGrams} and a repurposed version of \textit{Metroid}, thereby generalizing to beyond platformers.
This work thus contributes:
\begin{itemize}
 \item a controllable combinational creativity framework for blending games in terms of both levels and mechanics 
 \item implementations of the framework using both GMVAEs and CVAEs 
 \item to our knowledge, the first use of blended jump agents \item a novel CGMVAE model combining GMVAEs and CVAEs
\end{itemize}

\vspace{-0.75em}
\section{Related Work}
While PCGML \cite{summerville2017procedural} methods typically focus on modeling individual games, a body of work has emerged which instead recombines models from different games to generate new forms of content and explore design spaces across multiple games. Such works have recently been coined as methods for PCG via Knowledge Transformation \cite{sarkar2023pcgkt} and include game generation by recombining learned game graphs \cite{guzdial2021conceptual} and learning mappings between different platformers \cite{snodgrass2016approach}, to name a few.
A specific focus has been game blending i.e., generating new games by blending the levels and/or mechanics of two or more games, introduced in \cite{gow2015towards} where specifications of \textit{Frogger} and \textit{Zelda} were blended manually.
Several works have since implemented blending using variational autoencoders (VAEs) \cite{kingma2013autoencoding} which learn continuous latent representations that span all input games and can be sampled and searched for blended content, having been used for blending platformers \cite{sarkar2019blending,sarkar2020exploring} as well as platformers with dungeons \cite{sarkar2021dungeon}. These works however learn a fixed blend space without readily allowing the specification of different blend combinations such as blending a subset of the input games and in specific proportions. In this work, we implement controllable blending using Gaussian Mixture (GMVAE) and conditional VAEs (CVAE). GMVAEs \cite{dilokthanakul2016deep,shu2016gaussian} use a mixture of Gaussians as the prior distribution for the latent space unlike regular VAEs which use a standard Gaussian. This enables GMVAEs to cluster data in an unsupervised manner, where each GM component encodes a specific cluster and can be used to generate data belonging to it. Previously, GMVAEs have been used to discover clusters of platformer levels \cite{yang2020game}. We instead use supervised GMVAEs since we want each GM component to cluster levels of one of the input games and we know which game each level belongs to. 
\new{To our knowledge, this is the first use of a supervised GMVAE in a games context, though prior works have applied supervised \cite{cao2021open} and semi-supervised GMVAEs \cite{abdulaziz2022semi} outside of games.}
CVAEs \cite{sohn2015learning,yan2015conditional} add supervision to regular VAEs through labeling. 
Prior works \cite{sarkar2020conditional, sarkar2021dungeon} have used CVAEs to control blending using multi-hot labels that specify which games to include in the blend. We additionally apply \new{continuous} labels to control the relative ratios of each game.

Game blending is a conceptual blending \cite{fauconnier_conceptual_1998} method, falling under combinational creativity \cite{boden2004creative}, the branch of creativity concerned with generating new concepts by recombining existing ones. A prior method for generating new games similar to our approach is conceptual expansion \cite{guzdial2021conceptual}. Our works differ in that their system does automated game recombination and uses game graphs and video input \new{whereas} we generate specific combinations of games and use VAE latent representations and text-based input. In being a creative ML-based method for game design, our work can also be viewed as a GDCML (Game Design via Creative Machine Learning) approach \cite{sarkar2020towards}. 

Few previous works have attempted blending gameplay. 
Prior work \cite{cooper2022constraint} has demonstrated the use of a constraint-based method for blending the reachability rules of different games but the only prior instance of blending jumps can be found in \cite{summerville2020extracting} where blended jumps are extracted from paths in generated blended levels. We blend jumps of the original games directly using jump models from \cite{summerville2017mechanics} which learned hybrid automata to describe jumps of NES-era platformers.

\section{Framework}
We discuss our overall methodology in two sections. In this section, we introduce the framework and discuss its implementation via GMVAEs and CVAEs and its use for platformer level and jump blending. In the section after, we will discuss the CGMVAE and CCVAE architectures and how we use them to generate whole blended levels and extend to dungeons. The framework (Figure \ref{XFIGURElcgd}) consists of:
\begin{enumerate}
    \item \textit{Data} comprised of levels from $k$ different games
    \item A \textit{model} that learns a representation that enables working with different combinations of the $k$ games
    \item A set of $k$ \textit{weights} to specify how each game should be combined in the final blend
    \item A \textit{blended design space} obtained by blending the $k$ games using the weights
    \item A \textit{constraints module} that checks if outputs satisfy constraints such as playability.
\end{enumerate}

A dataset of $k$ games defines the full blending possibility space and is used to train a latent variable model capable of learning a disentangled\footnote{We use \textit{disentangled} in the literal sense \new{(i.e., the model learns separate representations for each game)} and not to refer to the specific technique of making latent dimensions learn independent factors of variation.} representation of the games. 
We use $k$-component GMVAEs and CVAEs with $k$-dimensional labels which allow manipulating each game via a separate GM component and label dimension respectively 
(note that this could be any model that allows reasoning about the $k$ subsets of data separately, such as a conditional GAN \cite{mirza2014conditional}).
To blend the games, we use $k$ weights defining the desired proportion of each game. Applying these gives us the final blended space with the $k$ games blended in the specified proportions. For the GMVAE, we linearly combine the $k$ GM components to obtain a new blended distribution. For the CVAE, we use the $k$ weights as the $k$-dimensional label when generating a level. The final blended design space is then sampled to produce the blended outputs.
Lastly, a module checks if the sampled output satisfies certain constraints. Here, the constraint is that outputs must be playable by blended agents\new{,} however this is a general module not restricted to playtesting e.g., it could be a module that uses evolutionary algorithms to search the design space for specific content.
Borrowing linear algebra vocabulary, we can consider the original $k$ games as the \textit{basis games} for blending and each blended game is a \textit{linear combination} of these basis games, specified by the blend weights. The set of all such linear combinations is the \textit{span} of these basis games. 

\vspace{-0.75em}
\subsection{Level Data}
Training data consisted of levels of \textit{Super Mario Bros. (SMB)}, \textit{Kid Icarus (KI)}, \textit{Mega Man (MM)} and \textit{Metroid (Met)}, taken from the Video Game Level Corpus (VGLC) \cite{summerville2016vglc}. 
VGLC levels use a text format with unique characters mapping to specific tiles. To show blending, we used a distinct mapping for each game, modifying the original tiles when games shared the same tile characters (e.g., enemies in both SMB and Met use `E'). We extracted 15x16 non-overlapping segments from each level based on the dimensions of playable areas in MM and Met where horizontal and vertical sections are 15 and 16 tiles high and wide respectively. KI levels are also 16 tiles wide. The 14-tile high SMB levels were padded with a row of background tiles on top. For Met, we filtered out entirely solid segments.
We extracted 172, 80, 143 and 435 segments for SMB, KI, MM and Met respectively and upsampled the non-Met segments to obtain a total of 1740 segments. 
Each segment had a length-4 one-hot label indicating which game it belonged to, with \clabel{1000}, \clabel{0100}, \clabel{0010} and \clabel{0001} indicating SMB, KI, MM and Met respectively.

\XFIGUREgmvaeboth

\vspace{-0.75em}
\subsection{Gaussian Mixture VAEs}
VAEs \cite{kingma2013autoencoding} are deep latent variable models consisting of encoder and decoder networks. For a $z$-dimensional latent space, for each input training instance, the encoder outputs $z$ pairs of means and variances which parameterize the latent distribution. A $z$-dimensional latent vector sampled from this distribution is then forwarded through the decoder to obtain outputs. Training is done by minimizing a loss function composed of---1) the reconstruction error between encoder inputs and decoder outputs and 2) the KL-divergence between the latent distribution and a tractable prior (typically standard Gaussian). 
Rather than use a standard Gaussian as the prior for the latent distribution, GMVAEs \cite{dilokthanakul2016deep,shu2016gaussian} use a mixture of Gaussians. For a mixture of $k$ Gaussians, this is done via: 1) using a $k$-dimension one-hot label indicating the component for an input, 2) training an additional 
network that maps a one-hot label to pairs of means and variances that parameterize the 1 out of $k$ components
the input belongs to and 3) modifying the 2nd loss term above to instead compute KL-divergence between the latent and component distributions. 
GMVAEs are typically unsupervised with a label-assigning network used to learn the one-hot labels. This was used in \cite{yang2020game} for discovering clusters of platformer levels. For blending however, we want each component to encode the segments of a specific game and since we know which game each segment belongs to, we use a supervised GMVAE, manually supplying one-hot labels for each segment, making the label-assigning network unnecessary and simplifying training, as shown in Figure \ref{XFIGUREgmvaeboth}. 
Thus, a $k$-component supervised GMVAE trained on levels from $k$ games results in each component encoding the levels for one game. Each game is thus modeled by a separate Gaussian distribution parametrized by a mean and a variance learned through training. For generating a blended game, we accept $k$ weights and model the blended game as a new distribution defined by the linear combination of the $k$ GMs, as specified by the weights.
For $k=4$ as in this work, after training we obtain 4 GM components with means and variances given by: $M = [\mu_1, \mu_2, \mu_3, \mu_4]$ and $V = [\sigma^2_1, \sigma^2_2, \sigma^2_3, \sigma^2_4]$. We then accept a weight vector $W = [w_1, w_2, w_3, w_4]$ indicating the weight of each GM desired in the new blended game. We then define a new blended game distribution as the linear combination of the 4 GMs with a mean given by the inner product $\langle M, W \rangle$ and variance given by the inner product $\langle V, W^2 \rangle$. 
This new distribution is the blended design space from the framework and is sampled to generate levels for the new blended game.





\vspace{-0.75em}
\subsection{Conditional VAEs}
Another approach for blending different combinations of games explored in past works involves conditional VAEs (CVAE) which are supervised VAEs that use labeled inputs. The encoder and decoder learn to use labels to encode inputs and produce outputs respectively. New outputs are generated by sampling a latent vector, concatenating the desired label and forwarding it through the decoder. Blending games using CVAEs involves using one-hot labels for training but multi-hot labels for generation e.g., \clabel{1011} indicates a blend of SMB/MM/Met without KI. Prior works have used CVAEs to blend games using multi-hot labels despite being trained only using one-hot labels. However, these labels only allow a game to be included or excluded from the blend. Since our goal is to generate arbitrary blends of games, we test if CVAEs trained using the same one-hot labels can work with arbitrary labels i.e., if similar blend weights used to linearly combine the learned components of the GMVAE, could be used by the CVAE to produce blended games that combine the games in accordance with the weights.
After training the CVAE on the same data and labels as the GMVAE, we generate blended levels by: 1) sampling vectors from the CVAE latent space 2) accepting $k$ blend weights as before but using them as the $k$-element label 3) concatenating it to each sampled vector and 4) forwarding the concatenated vectors through the decoder to obtain the output blended levels.

\vspace{-0.5em}
\subsection{Blending Jumps}
Prior works \cite{sarkar2020exploring, sarkar2021generating} have evaluated playability of blended platformer levels using separate A* agents for each game. By seeing which game's agent can play through a blended level, we can conclude if a blended level is e.g., more Mario-like or Metroid-like. However, since our goal is generating new blended games, we blend gameplay in addition to blending levels. 
We define a blended jump model as a linear combination of the parameters of the original jump models, using the same weights used to blend levels. 
Thus, the jump for a blended game is a linear combination of the jumps of the original games. 
For jump models, we use the models from prior work  \cite{summerville2017mechanics} that learned hybrid automata for describing the jump physics of several NES platformers, including the four we use. The automata consist of parameters defining the jump arcs and vary depending on the game. We obtain blended jump models by linearly combining these parameters using the blend weights. 
The automata can be used to derive jump arcs for the respective games. These jumps, when represented as lists of (x,y) coordinates, can in turn be used to learn the impulse and gravity parameters that define the jump, as described in \cite{summerville2020extracting}. This has two specific benefits---1) jump arcs derived via blending can be used with the Summerville tile-based A* agent \cite{summerville2016mariostring} to quickly test if they can be used to traverse blended levels and 2) in the future, the impulse and gravity parameters could be used to define the mechanics for a player controller so that the blended levels can be played in an interactive environment.

\vspace{-0.5em}
\subsection{Experiments}
All models were implemented using PyTorch \cite{paszke2019pytorch}.
The GMVAE consists of an encoder, decoder and prior-assigning network.
The encoder and decoder consisted of 4 and 3 fully-connected layers respectively, with both using ReLU activation. The last encoder layer was split into 2 parts, each having 1 fully-connected layer with the latter additionally using softplus activation. These two layers output the means and variances of the latent distribution respectively. The prior assigning network consisted of 2 independent sub-networks each having 1 fully-connected layer with the second additionally using Softplus activation. The two networks output the component means and variances respectively.
Training was done for 1000 epochs using the Adam optimizer with an initial learning rate of 0.001, decayed by 0.1 every time training plateaued for 50 epochs.
For the CVAE, the encoder and decoder consisted of 4 and 3 fully-connected layers respectively, both using ReLU activation. Training lasted 10000 epochs using Adam with an initial learning rate of 0.001, set to decay by 0.1 every 2500 epochs. The weight on the KL term of the loss function was annealed from 0 to 1 for the first 2500 epochs. Hyperparameters were determined based on prior works using these models and through experimentation. 
For evaluation, we conducted a 3-part study measuring 1) how accurately the output of the blend models adhered to the blend weights, 2) playability of the blended levels using the blended agents and 3) comparing the tile patterns and content in the generated blended levels with those in the original levels. 
Additionally, for all evaluations, we used two types of blend weights---1) binary weights in the form of length-4 one-hot and multi-hot vectors indicating single-game and multi-game blends respectively and 2) fractional weights consisting of length-4 floating-point vectors, representing any arbitrary combination of games. For binary weights, we evaluate all possible length-4 binary vectors except \clabel{0000} since it is unclear how to reason about a blend containing none of the games, giving us $2^4-1=15$ vectors. For fractional weights, we define 4 vectors to test against to cover a range of blend combinations: \clabel{0.5, 0.3, 0.2, 0.0}, \clabel{0.1, 0.1, 0.1, 0.7}, \clabel{0.1, 0.6, 0.2, 0.1} and \clabel{0.0, 0.2, 0.3, 0.5}.
For each of the CVAE and GMVAE, we tested 4 different latent dimensionalities of 8, 16, 32 and 64 but found none to be clearly preferable across all experiments, with different sizes producing best results for different metrics.
For clarity and space, instead of presenting results for all 8 models, we present results from the GMVAE-32 and CVAE-64 models as they performed the best on the blend accuracy evaluation which is the most relevant for the  main focus of this work i.e., controllable blending of games.

\XTABLEclassboth

\subsubsection{Blend Accuracy}
For evaluating how well the outputs of the blended models adhere to the blend weights, we used scikit-learn \cite{pedregosa2011scikit} to train a random forest classifier on the segments using the game of that segment as the class label. We achieved a 98.19\% accuracy on an 80-20 train-test split. The parameters for the classifier were determined via grid search. 
This classifier approach for evaluating blending has been used in prior CVAE works and acts as a proxy since we lack ground-truth blended levels to compare generated output. The expectation is that when using single game weights (e.g., \clabel{0,0,0,1}, \clabel{1,0,0,0}), classifier predictions for a game will be high when its bit is set to 1 and low when set to 0. Similarly, when using blended weights (e.g., \clabel{1,1,0,0}, \clabel{0,1,0,1}), the predictions should be more spread out across the games whose bit is set to 1. When using fractional weights (e.g., \clabel{0.5, 0.2, 0.3, 0.0}), the predictions should be spread out in accordance with these weights.
For classifying binary weights, for each model, we sampled 1000 latent vectors for each possible 4-digit binary weight, leaving out \clabel{0,0,0,0}. We applied the classifier on each generated segment and tracked the percentage of times each of the 4 games was predicted, per 4-digit weight. To determine how well the outputs matched a given weight, for each weight, we computed a score based on the following formula:
$S = \sum_{i=1}^{4} (w_i*f - p_i)^2 $
where $w_i$ is the $i$th weight element indicating the desired weight of the $i$th game, $f$ is a factor equal to 100 divided by the number of ones in the weight and $p_i$ is the percentage of the 1000 segments for which the classifier predicted the $i$th game. The lower the score, the better the model matched outputs with the given weights with 0 representing a perfect match, e.g., if the weights are \clabel{1,0,1,0}, the factor $f$ is 100/2=50 and a perfect score of 0 is achieved when the classifier predicts games 1 and 3 50\% of the time each.
For fractional weights, we generated 1000 latents per weight per model and computed scores similarly, except here $f$ is set to a constant 100.
Results are given in Table \ref{XTABLEclassboth}.
For both binary and fractional weights, the GMVAE does better than the CVAE for all latent dimensions except for 64 where the CVAE does better. Looking at the latent sizes, for the GMVAE and CVAE, 32 and 64 dimensions respectively give the lowest score when combining binary and fractional weights and thus for the remainder of our evaluations, we will only present and discuss results obtained using the 32-dimensional GMVAE and 64-dimensional CVAE.
Full classification results are shown in Table \ref{XTABLEclassfull}. We see that in most cases, the game predicted most often is one whose bit is set to 1 in the binary case. When using fractional weights, the amounts that levels are classified as different games corresponds to the value of the weights, though very loosely.
Interestingly, the CVAE is able to reason with fractional weights despite having only been trained using one-hot binary weights as labels. Overall, this suggests that both GMVAE and CVAE models are able to generate output that corresponds to the desired blend weights.
These results give some intuition about why the GMVAE outperforms the CVAE in Table \ref{XTABLEclassboth}\new{,} namely that the GMVAE is better able to blend the individual games in accordance with the weights while the CVAE is more prone to pushing the blend towards the game(s) with highest weight, \new{as shown (in Table \ref{XTABLEclassfull}), by how for most multi-game blends, the CVAE scores higher (worse) than the GMVAE, usually due to putting too much weight on one of the games whose bit is set to 1.
This may be due to how these models perform blending. GMVAEs learn separate distributions per game which are then combined using the weights where as CVAEs learn one distribution spanning all games and blends are produced by applying labels (i.e., weights) to samples from this distribution. If this single distribution approximates some games better than others, the labels may be insufficient in producing accurate blends.}

\XTABLEclassfull

\XFIGUREjump

\subsubsection{Playability}
We evaluated playability of generated blended levels using jump arcs obtained by blending the automata models using given blend weights and the Summerville tile-based A* agent \cite{summerville2015mcmcts}. The original VGLC version worked with solid and non-solid tiles and was updated in  \cite{summerville2020extracting} to work with 4 affordances \cite{bentley2019videogame}: solids, hazards, passables and climbables. 
We further updated the agent to find start and goal tiles in a segment. A segment is then playable if the agent can find a path from start to goal. If no path or goal is found, it is unplayable. The jumps used by the blend agent are derived from blended automata models which in turn are obtained by blending the hybrid automata models of the original 4 games using the blend weights. Details about these automata and jump extraction can be found in \cite{summerville2017mechanics}. Blended jump arcs are shown in Figure \ref{XFIGUREjump}. Since the blend agent simulates an agent that can play blends of all 4 games, it knows the tile-to-affordance mappings of all 4 games.
For each model, we sampled 1000 segments for each possible 4-digit binary weight and computed the percentage that were playable by the blended agent obtained by blending the original jump models according to the corresponding weight. Since games can progress both horizontally and vertically, for each segment, the agent looks for both a valid horizontal and vertical path. If either one is found, the segment is playable. To simplify computations, we only look for left-to-right and bottom-to-top paths. Thus, our evaluations underestimate playability since e.g., MM and \new{Met} segments that progress downward but do not have an upward moving path would be deemed unplayable. 
Full playability results for the blended agent on binary and fractional-weighted blends are shown in Tables \ref{XTABLEblendagent} and \ref{XTABLEblendagentfrac}. Table \ref{XTABLEblendagentfrac} also shows how well the original agents did in the blended levels to depict that performance roughly corresponds to the game's weight in the blend e.g., for \clabel{0.5, 0.3, 0.2, 0}, the SMB agent completes more levels than the KI agent which completes more than the MM agent which does better than the Met agent. We show the performance of the original agents to show that all models generate levels in accordance with the weights and not to compare playability of the blend agent with the original agents since such comparisons are not useful. Essentially, there are two broad ways to compare. First, we can assume that the original agent for a game should not reason about the tiles of another e.g., an SMB agent should not know to avoid MM spikes. But if only the blend agent knows all tiles across all games, then it will trivially do the best since in most blended levels, the original agents would face tiles that are undefined for them and they would not be able to find paths. Thus, playability reduces to tile knowledge.
Second, we can instead assume that all agents work at the affordance level, agnostic of game-specific tiles. Here, playability is completely determined by an agent's jumpsize. Higher the jump, higher the playability when all agents work with the same tile information. Since the blend agent combines the jumps of all games, its jump by definition can never be bigger than that of Metroid. Thus, playability reduces to jumpsize. Hence, in either case, such comparisons are uninformative. Overall, we observe satisfactory playability of 61.97\% and 65.58\% for the GMVAE, averaged across binary and fractional weights respectively and 70.14\% and 68.03\% for the CVAE.

\XTABLEblendagent

\XTABLEblendagentfrac

\XFIGUREexamples

\subsubsection{Tile Metrics}
To compare the content of generated blended levels with those in the original levels, we used the Tile Pattern KL-Divergence (TPKLDiv) \cite{lucas2019tile} metric which measures the similarity between two sets of levels in terms of the KL-divergence between their tile pattern distributions. For each model and set of blend weights, we generate 1000 levels and compare the average TPKLDiv between them and each set of original levels, separately for each of the 4 games, with the values averaged over 2x2, 3x3 and 4x4 patterns. We expect lower values when comparing the original levels of a game with blended levels produced by weights that include that game, and higher values when they do not, e.g., using \clabel{1,1,0,0} should lead to lower TPKLDiv values compared to SMB and KI than with MM and Met. Full results are in Table \ref{XTABLEtpklbinfull} and are true to this expectation, giving further support that the models can produce blends in accordance with the weights.

\XTABLEtpklbinfull 

\subsection{Visual Inspection}
Figure \ref{XFIGUREexamples} shows example blended segments. 
\new{Note that a sampled segment may not appear blended since weights apply to whole distributions and not individual segments. Thus it is possible to sample segments that contain content from only 1 game}, highlighting the need to use more sophisticated methods beyond just random sampling. That said, these examples depict the games being blended. SMB can be identified by its pipes and goombas, KI by its blue platforms, red hazards, brown tiles and doors, MM via orange tiles, spikes and ladders and Metroid via dark blue tiles, orange lava and metroid creatures.

\section{Generating and Blending Whole Levels}
In this section, we describe our approach for extending the LCGD framework in 2 ways---1) generating whole levels and 2) blending levels from dungeon-based games.
For this, we combine the GMVAE and CVAE to introduce a novel Conditional Gaussian Mixture VAE (CGMVAE) architecture which uses a supervised GM prior for the VAE latent space thus still enabling games to be encoded as GM components but now, we also condition the encoder/decoder with labels indicating segment orientation i.e., the direction(s) in which movement is possible across the segment. Directional labeling was introduced in prior work \cite{sarkar2021dungeon} for generating whole levels. Using labels for progression enables generating segments that have desired orientations and can be reliably connected together to form whole levels. For dungeons, such labels can control the direction(s) in which rooms have doors/openings and enable generating dungeons with appropriately interconnected rooms. This in turn lets us use the CGMVAE to apply the LCGD framework on dungeon-based games. We compare this with a CVAE where a label controlling direction is concatenated to the label controlling blending. To distinguish it from the prior CVAEs, we refer to it as CCVAE to indicate that the labels here control both blend ratio and directionality.

\XFIGUREcgmvae

\vspace{-0.75em}
\subsection{Model and Architecture}
The architectures and training for both CGMVAE and CCVAE were identical to the GMVAE and CVAE but augmented by the length-4 multi-hot labels. For both models, these labels were concatenated to both the encoder and decoder inputs. The CGMVAE architecture is shown in Figure \ref{XFIGUREcgmvae}.
The reasoning for this labeling is that whole levels can be considered to be composed of segments that are connected together so as to enable progressing through them. Platformer and dungeon levels can be viewed as made up of discrete segments and rooms respectively. 
For the CCVAE, we concatenated the new directional label to the game label used with the CVAE models.

\XTABLEdirlabels

\vspace{-0.75em}
\subsection{Level Data}
The training data included levels from both platformers and dungeon-based games. The platformer models were trained using the same data as before, with directional labels added to the game labels. Directional labels were length-4 multi-hot vectors with elements corresponding to up, down, left and right respectively and set to 1/0 to indicate if progression in the corresponding direction was/was not possible e.g., \clabel{1,1,0,0} indicates a segment where progress can be made upward or downward but not left or right. The label for each segment was determined manually via visual inspection and indicated the directions in which a segment could be entered or exited by a player. For Metroid segments, directions having a door were considered open.
A challenge for training on dungeon-based games is the lack of data in the VGLC where the only such game with text-based level data is \textit{The Legend of Zelda}. Thus, to blend multiple such games, we also used levels from \textit{DungeonGrams} (DGG), a roguelike dungeon crawler developed for prior research \cite{biemer2021gram}. We also leveraged the previously used Metroid levels. While Metroid is a platformer, its sprawling, interconnected world can be seen as a dungeon from a top-down perspective.
We trained on 15x16 segments from each game. Since Zelda and DGG levels are 11x16, we padded each by duplicating the outermost rows. We also added vertically and horizontally flipped versions of each Zelda room, if not already present. We obtained 502 Zelda rooms, 522 DGG levels and 435 Metroid segments but upsampled Zelda and Metroid to get a total of 1566 segments. Example segments and directional labels are shown in Table \ref{XTABLEdirlabels}. 

\vspace{-0.75em}
\subsection{Layout Algorithm}
To arrange sampled segments/rooms into whole levels, we use the layout algorithm from \cite{sarkar2021dungeon}. This starts with an initial segment location closed in all four directions. In each iteration, we randomly select a closed side of the current location and place a new location next to that side if there is none. The closed side just selected is then labeled open to connect the new location to the prior one. This is repeated until a desired number of interconnected locations are generated. Generating a platformer layout is simpler since movement is overall left-to-right, with/without vertical sections in between. For this, we randomly set an upward or rightward direction for the initial segment and set directions for subsequent segments such that progress is possible from start to finish (e.g., segment following a rightward/upward segment should be open on the left/bottom respectively). There are many ways for arranging segments into whole levels and in the context of the framework, the layout algorithm could be viewed as part of the constraints module.
To produce the final level, we loop through each location, determine the label based on the sides that are open/closed, sample a latent vector, concatenate the label and forward it through the decoder to generate the segment/room. For the CGMVAE, the directional label is concatenated as in Figure \ref{XFIGUREcgmvae}. For the CCVAE, both the game blend label and the directional label are concatenated to the sampled latent.

\vspace{-0.75em}
\subsection{Experiments}
For evaluations, we tested that 1) the new directional labels do not prevent the new models from blending in accordance with the blend weights and 2) the generated outputs are open/closed in accordance with the directional labels. 

\subsubsection{Blend Accuracy}
We performed an identical classifier-based evaluation as for the prior models. We re-used the random forest classifier trained previously for the platformer models and trained a new random forest classifier for the dungeon models. We achieved a 98.8\% accuracy on a 80-20 train-test split for the new classifier. Results for platformers and dungeon games are given in Tables \ref{XTABLEclassfullplt} and \ref{XTABLEclassfulldg} respectively. Similar to the prior models, for both platformers and dungeon games, the game classifications using the new models are in accordance with the blend weights in a majority of cases. That is, the game predicted most frequently by the classifier is one whose bit is set to 1 in the binary case. The models do less well for the fractional weights with most predictions being for the game with the highest weight in only 5 out of 8 and 4 out of 6 model-weight pairing for platformers and dungeons respectively with \new{CCVAE} doing worse than \new{CGMVAE} for dungeons for both types of weights. To compare the two models, we performed the same evaluation as done for the prior models (i.e., the one in Table \ref{XTABLEclassboth}) using the same approach described previously. Results are shown in Table \ref{XTABLEcgmccclass} and confirm that in all but one case, the \new{CGMVAE} outperforms the \new{CCVAE}, especially when using binary weights. Note that these results are consistent with the corresponding evaluation for the initial GMVAE and CVAE models, thus suggesting that the additional labels for controlling directionality do not negatively impact the ability of these models to blend games and that like the GMVAE outperforming the CVAE, here the CGMVAE outperforms the CCVAE, with the CCVAE only doing better for fractional weights in the platformer case. 
\new{This worse performance may stem from the difference in how they enable blending compared to CGMVAE, as previously noted.}

\XTABLEclassfullplt

\XTABLEclassfulldg

\XTABLEcgmccclass

\subsubsection{Directional Accuracy}
For this, we wanted to test if generated segments had directionality/orientation true to the labels used to generate them. Similar to the blend label evaluation, we trained a random forest classifier on the segments of the three games to predict their directional labels, obtaining 97.13\% accuracy using an 80-20 train-test split. For our evaluation, for each generated segment, we compared the label predicted by this classifier with the directional label used to condition its generation. Ideally, we would like the predicted label to exactly match the label used for generation i.e., the generated segment would be open/closed in the precise directions indicated by the label. However, as noted in \cite{sarkar2021dungeon}, it is sufficient for the generated segment to match in only the required open directions to ensure playability, irrespective of whether it is open/closed in the desired closed directions. We only really want to avoid the opposite case i.e., the generated segment being closed off in a desired open direction. Borrowing terminology from \cite{sarkar2021dungeon}, we use the notion of exact and admissible matches where the former refers to cases where the predicted label exactly matches the conditioning label and the latter to situations where the bits set to 1 in the conditioning label are also 1 in the predicted label, regardless of the value of the bits set to 0 in the conditioning label. Thus, by definition, an exact match is also admissible. For our evaluation, we sampled 1000 latent vectors for each blend weight and conditioned the generation of each using all 15 directional labels. We computed the mean percentage of exact, admissible and inadmissible matches for each weight, averaged across the 15 labels. 
Results are shown in Table \ref{XTABLEdirsummboth}. While exact matches are not high for any model, the percentage of admissible matches (which includes exact matches) is about 80\% in most cases, with the number being lower for the binary platformer models, albeit still much higher than the inadmissible matches. Thus, using the hybrid models, we can both reliably obtain desired blended distributions and also control the direction of segments sampled from them. 

\XTABLEdirsummboth

\subsubsection{Tile Metrics}
We did a TPKLDiv-based evaluation for the new models with results shown in Tables \ref{XTABLEtpklbinfullplt} and \ref{XTABLEtpklbinfulldg}. In \new{essentially} all cases, games with the highest/joint highest and lowest/joint lowest weight have the smallest/highest TPKLDiv to the corresponding original game respectively. This suggests that, in terms of tile patterns, the blended distributions are closer/farther to the original games in accordance with the blend weights, thus indicating that incorporating directional conditioning to the GMVAE and CVAE does not negatively impact blending, as also suggested by the blending evaluation.

\XTABLEtpklbinfullplt

\XTABLEtpklbinfulldg

\subsection{Visual Inspection}
Sample whole levels are shown in Figures \ref{XFIGUREplatformeggm}-\ref{XFIGUREdungeonegcc}, generated using the layout algorithm described previously with equal weights for each game i.e., 1/4 each for platformers and 1/3 each for dungeons. While these are \new{manually selected}, they are representative of the merits of the GMVAE over CVAE in better respecting the blend weights as seen in Figures \ref{XFIGUREplatformeggm} and \ref{XFIGUREdungeoneggm} which blend content from all platformers and dungeon games while there is little Mario and DGG content (brown tiles) in Figures \ref{XFIGUREplatformegcc} and \ref{XFIGUREdungeonegcc} respectively. This reflects how the game classifications for the CGMVAE are more spread out in accordance with the weights compared to the CCVAE as shown in Table \ref{XTABLEcgmccclass}.
\new{These visualizations show some amount of noise in generated levels which is expected when blending games. Future work could reduce noise by testing different architectures/training. Our framework is not specific to VAEs so we could try models such as CGANs \cite{mirza2014conditional} and GMGANs \cite{ben2018gaussian}. We could also conduct user studies to test how preferences and qualitative evaluations match with quantitative evaluations e.g., a model with better metrics might produce levels that a user finds less aesthetically pleasing and vice-versa.}

\XFIGUREplatformeggm

\XFIGUREplatformegcc

\XFIGUREdungeoneggm

\XFIGUREdungeonegcc

\vspace{-0.25em}
\new{\section{Discussion and Limitations}}
\new{In this section we discuss limitations of this work and some future directions for more thoroughly investigating the nature of blending. 
We note the classifier-based approach is a proxy for the lack of ground-truth blended levels and recognize its issues, e.g., given a blend label \clabel{1001}, a perfect classification score of 0 can be achieved by a generator outputting a Mario-only level or a Metroid-only level 50\% of the time each. While problematic in a vacuum, our models were trained without any signal from a classifier and thus do not learn this adversarial behavior. Combined with the other experiments, the classifier evaluations lend support that the models respect the blend weights. A limitation we do observe is that some games can be poorly represented in the blends, particularly Metroid for platformers and Zelda for dungeons. This may be due to Metroid being the most structurally different from other platformers and thus its latent encodings end up further from those of the other games. When blending the learned game distributions using equal weight, it is possible that the final blend is furthest from the Metroid distribution, leading to fewer Metroid-like segments being sampled. Similarly, for dungeon blends, Zelda with its discrete, self-contained rooms is different than both DGG and Metroid with their more open structures. This could be fixed by scaling the blend weights for these outlier games so that the final blend is closer to the desired combination.
Further, we want to more precisely investigate the nature of blending, both in terms of how the levels are blended across games as well as the interaction of level and mechanic blending. Based on evaluations, visual inspection and our intuition, in blending, the models seem to generate game-specific tiles in game-agnostic semantically-equivalent locations in terms of affordances, which is a reasonable analog to how a designer might blend such games. Thus, in addition to our TPKLDiv evaluations, we could evaluate the underlying affordances and compute affordance-based KL-Divergence as in \cite{sarkar2022tile}.
We could also decouple the level and mechanic blends i.e., test different jump blends on a given level blend and vice-versa, which could enable mixing and matching different level and mechanic blends to produce an even greater variety of games. Relatedly, our approach could benefit from additional methods of controlling outputs sampled from the blended distribution since currently one can only sample at random. While some control is provided by the CGMVAE and CCVAE  in terms of orientation, additional means of control could leverage prior work that combines similar models with evolutionary algorithms to search for specific content \cite{bontrager2018deepmasterprints} or evolve an archive of playable levels \cite{gravina2019procedural} based on specified behavioral features. A method for more reliably sampling playable levels is also needed. For this, we could incorporate into the constraints module an A* agent that repairs unplayable levels as in \cite{cooper2020pathfinding}. 
}

\vspace{-0.5em}
\section{Conclusion and Future Work}
We introduced \textit{latent combinational game design}, a framework for controllably blending games, demonstrated it with a supervised GMVAE and a CVAE and 
proposed a new CGMVAE architecture which combines the two models to generate whole blended levels.
Our future goal is to build a system where the blended games can be played. 
We also wish to blend mechanics beyond jumps by incorporating affordances and try alternatives besides A* agents for the constraints module, such as an ASP solver.
\new{We could also try learning mechanics for a given blended domain by, for example, training an RL agent to learn actions given a set of levels and tile affordances.}
Future work could also use evolution to search the blended space instead of sampling randomly.
Finally, the framework could be used as a combinational creativity approach for artistic domains in general e.g., applied on a dataset of different styles of paintings to produce a custom blend of art styles.
\vspace{-0.75em}